# Sentence Compression as Tree Transduction

**Trevor Cohn**                                              TCOHN@INF.ED.AC.UK

**Mirella Lapata**                                           MLAP@INF.ED.AC.UK
*School of Informatics*
*University of Edinburgh*
*10 Crichton Street Edinburgh EH8 10AB, UK*

## Abstract

This paper presents a tree-to-tree transduction method for sentence compression. Our model is based on synchronous tree substitution grammar, a formalism that allows local distortion of the tree topology and can thus naturally capture structural mismatches. We describe an algorithm for decoding in this framework and show how the model can be trained discriminatively within a large margin framework. Experimental results on sentence compression bring significant improvements over a state-of-the-art model.

## 1. Introduction

Recent years have witnessed increasing interest in text-to-text generation methods for many natural language processing applications, ranging from text summarisation to question answering and machine translation. At the heart of these methods lies the ability to perform rewriting operations. For instance, text simplification identifies which phrases or sentences in a document will pose reading difficulty for a given user and substitutes them with simpler alternatives (Carroll, Minnen, Pearce, Canning, Devlin, & Tait, 1999; Chandrasekar & Srinivas, 1996). In question answering, questions are often paraphrased in order to achieve more flexible matching with potential answers (Lin & Pantel, 2001; Hermjakob, Echihabi, & Marcu, 2002). Another example concerns reformulating written language so as to render it more natural sounding for speech synthesis applications (Kaji, Okamoto, & Kurohashi, 2004).

Sentence compression is perhaps one of the most popular text-to-text rewriting methods. The aim is to produce a summary of a single sentence that retains the most important information while remaining grammatical (Jing, 2000). The appeal of sentence compression lies in its potential for summarization and more generally for document compression, e.g., for displaying text on small screens such as mobile phones or PDAs (Vandeghinste & Pan, 2004). Much of the current work in the literature focuses on a simplified formulation of the compression task which does not allow any rewriting operations other than word deletion. Given an input *source* sentence of words $\mathbf{x} = x_1, x_2, \ldots, x_n$, a *target* compression $\mathbf{y}$ is formed by removing any subset of these words (Knight & Marcu, 2002).

Despite being restricted to word deletion, the compression task remains challenging from a modeling perspective. Figure 1 illustrates a source sentence and its target compression taken from one of the compression corpora used in our experiments (see Section 5 for details). In this case, a hypothetical compression system must apply a series of rewrite rules in order to obtain the target, e.g., delete the leaf nodes *exactly* and *and*, delete the subtrees *made it* and *which ones*, and merge the subtrees corresponding to *records* and *are involved*. More concretely, the system must have access to rules like those shown in Figure 2. The rules are displayed as a pair of tree fragments where the left fragment corresponds to the source and the right to the target. For instance, rule (1) states that a wh-noun phrase (WHNP) consisting of an adverb (RB) and a wh-pronoun (WP) (e.g., *exactly what*) can be rewritten as just a wh-pronoun (without the adverb). There are two things to note here. First, syntactic information plays an important role, since deletion decisions are not limited to individual words but often span larger constituents. Secondly, there can be a large number of compression rules of varying granularity and complexity (see rule (5) in Figure 2).

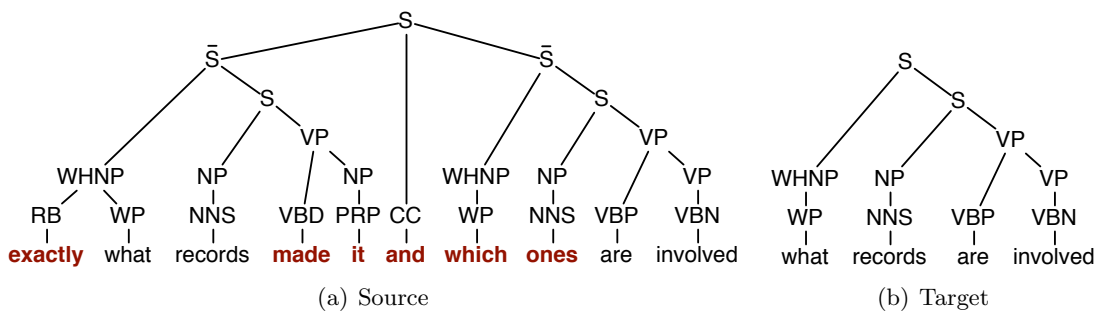

(a) Source　　　　　　　　　　　(b) Target

Figure 1: Example of sentence compression showing the source and target trees. The bold source nodes show the terminals that need to be removed to produce the target string.

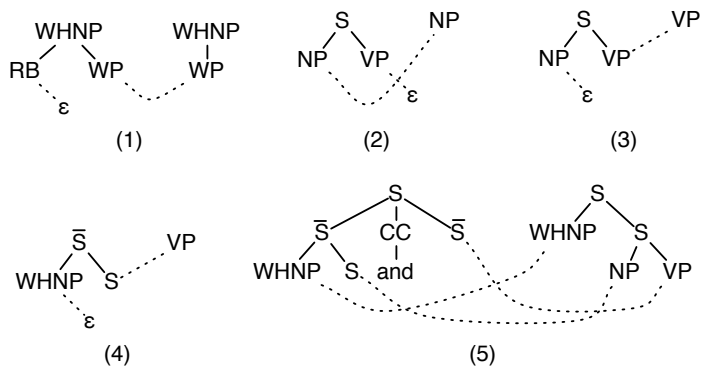

Figure 2: Example transduction rules, each displayed as a pair of tree fragments. The left (source) fragment is matched against a node in the source tree, and the matching part is then replaced by the right (target) fragment. Dotted lines denote variable correspondences, and $\epsilon$ denotes node deletion.

Previous solutions to the compression problem have been cast mostly in a supervised learning setting (for unsupervised methods see Clarke & Lapata, 2008; Hori & Furui, 2004; Turner & Charniak, 2005). Sentence compression is often modeled in a generative framework where the aim is to estimate the joint probability $P(\mathbf{x}, \mathbf{y})$ of source sentence $\mathbf{x}$ having the target compression $\mathbf{y}$ (Knight & Marcu, 2002; Turner & Charniak, 2005; Galley & McKeown, 2007). These approaches essentially learn rewrite rules similar to those shown in Figure 4 from a parsed parallel corpus and subsequently use them to find the best compression from the set of all possible compressions for a given sentence. Other approaches model compression discriminatively as subtree deletion (Riezler, King, Crouch, & Zaenen, 2003; Nguyen, Horiguchi, Shimazu, & Ho, 2004; McDonald, 2006).

Despite differences in formulation, existing models are specifically designed with sentence compression in mind and are not generally applicable to other tasks requiring more complex rewrite operations such as substitutions, insertions, or reordering. A common assumption underlying previous work is that the tree structures representing the source sentences and their target compressions are *isomorphic*, i.e., there exists an edge-preserving bijection between the nodes in the two trees. This assumption is valid for sentence compression but does not hold for other rewriting tasks. Consequently, sentence compression models are too restrictive; they cannot be readily adapted to other generation problems since they are not able to handle structural and lexical divergences. A related issue concerns the deletion operations themselves which often take place without considering the structure of the target compression (the goal is to generate a compressed string rather than the tree representing it). Without a syntax-based language model (Turner & Charniak, 2005) or an explicit generation mechanism that licenses tree transformations there is no guarantee that the compressions will have well-formed syntactic structures. And it will not be straightforward to process them for subsequent generation or analysis tasks.

In this paper we present a sentence compression model that is not deletion-specific but can account for ample rewrite operations and scales to other rewriting tasks. We formulate the compression problem as tree-to-tree rewriting using a synchronous grammar (with rules like those shown in Figure 2). Specifically, we adopt the *synchronous tree substitution grammar* (STSG) formalism (Eisner, 2003) which can model non-isomorphic tree structures while having efficient inference algorithms. We show how such a grammar can be induced from a parallel corpus and propose a discriminative model for the rewriting task which can be viewed as a weighted tree-to-tree transducer. Our learning framework makes use of the large margin algorithm put forward by Tschantaridis, Joachims, Hofmann, and Altun (2005) which efficiently learns a prediction function to minimize a given loss function. We also develop an appropriate algorithm that can be used in both training (i.e., learning the model weights) and decoding (i.e., finding the most plausible compression under the model). Beyond sentence compression, we hope that some of the work described here might be of relevance to other tasks involving structural matching (see the discussion in Section 8).

The remainder of this paper is structured as follows. Section 2 provides an overview of related work. Section 3 presents the STSG framework and the compression model we employ in our experiments. Section 5 discusses our experimental set-up and Section 6 presents our results. Discussion of future work concludes the paper.

## 2. Related Work

Synchronous context-free grammars (SCFGs, Aho & Ullman, 1969) are a generalization of the context-free grammar (CFG) formalism to simultaneously produce strings in two languages. They have been used extensively in syntax-based statistical MT. Examples include inversion transduction grammar (Wu, 1997), head transducers (Alshawi, Bangalore, & Douglas, 2000), hierarchical phrase-based translation (Chiang, 2007), and several variants of tree transducers (Yamada & Knight, 2001; Grael & Knight, 2004).

Sentence compression bears some resemblance to machine translation. Instead of translating from one language into another, we are translating long sentences into shorter ones within the same language. It is therefore not surprising that previous work has also adopted SCFGs for the compression task. Specifically, Knight and Marcu (2002) proposed a noisy-channel formulation of sentence compression. Their model consists of two components: a language model $P(\mathbf{y})$ whose role is to guarantee that the compression output is grammatical and a channel model $P(\mathbf{x}|\mathbf{y})$ capturing the probability that the source sentence $\mathbf{x}$ is an expansion of the target compression $\mathbf{y}$. Their decoding algorithm searches for the compression $\mathbf{y}$ which maximizes $P(\mathbf{y})P(\mathbf{x}|\mathbf{y})$. The channel model is a stochastic SCFG, the rules of which are extracted from a parsed parallel corpus and their weights estimated using maximum likelihood. Galley and McKeown (2007) show how to obtain improved SCFG probability estimates through Markovization. Turner and Charniak (2005) note that SCFG rules are not expressive enough to model structurally complicated compressions as they are restricted to trees of depth 1. They remedy this by supplying their synchronous grammar with a set of more general "special" rules. For example, they allow rules of the form $\langle\text{NP,NP}\rangle \rightarrow \langle[\text{NP NP}_{\boxed{1}} \text{ CC NP}_{\boxed{2}}], \text{NP}_{\boxed{1}}\rangle$ (boxed subscripts are added to distinguish between the two NPs).

Our own work formulates sentence compression in the framework of synchronous tree-substitution grammar (STSG, Eisner, 2003). STSG allows to describe non-isomorphic tree pairs (the grammar rules can comprise trees of arbitrary depth) and is thus suited to text-rewriting tasks which typically involve a number of local modifications to the input text. Especially if each modification can be described succinctly in terms of syntactic transformations, such as dropping an adjectival phrase or converting a passive verb phrase into active form. STSG is a restricted version of synchronous tree adjoining grammar (STAG, Shieber & Schabes, 1990) without an adjunction operation. STAG affords mild context sensitivity, however at increased cost of inference. SCFG and STSG are weakly equivalent, that is, their string languages are identical but they do not produce equivalent tree pairs. For example, in Figure 2, rules (1)–(4) can be expressed as SCFG rules, but rule (5) cannot because both the source and target fragments are two level trees. In fact it would be impossible to describe the trees in Figure 1 using a SCFG. Our grammar rules are therefore more general than those obtained by Knight and Marcu (2002) and can account for more elaborate tree divergences. Moreover, by adopting a more expressive grammar formalism, we can naturally model syntactically complex compressions without having to specify additional rules (as in Turner & Charniak, 2005).

A synchronous grammar will license a large number of compressions for a given source tree. Each grammar rule typically has a score from which the overall score of a compression $\mathbf{y}$ for sentence $\mathbf{x}$ can be derived. Previous work estimates these scores generatively as discussed above. We opt for a discriminative training procedure which allows for the incorporation of all manner of powerful features. We use the large margin technique proposed by Tsochantaridis et al. (2005). The framework is attractive in that it supports a configurable loss function, which describes the extent to which a predicted target tree differs from the reference tree. By devising suitable loss functions the model can be straightforwardly adapted to text rewriting tasks besides sentence compression.

McDonald (2006) also presents a sentence compression model that uses a discriminative large margin algorithm. The model has a rich feature set defined over compression bigrams including parts of speech, parse trees, and dependency information, without however making explicit use of a synchronous grammar. Decoding in this model amounts to finding the combination of bigrams that maximize a scoring function defined over adjacent words in the compression and the intervening words which were dropped. Our model differs from McDonald's in two important respects. First, we can capture more complex tree transformations that go beyond bigram deletion. Being tree-based, our decoding algorithm is better able to preserve the grammaticality of the compressed output. Second, the tree-based representation allows greater modeling flexibility, e.g., by defining a wide range of loss functions over the tree or its string yield. In contrast, McDonald can only define loss functions over the final compression.

Although the bulk of research on sentence compression relies on parallel corpora for modeling purposes, a few approaches use no training data at all or a small amount. An example is in the work of Hori and Furui (2004), who propose a model for automatically transcribed spoken text. Their method scores candidate compressions using a language model combined with a significance score (indicating whether a word is topical or not), and a score representing the speech recognizer's confidence in transcribing a given word correctly. Despite being conceptually simple and knowledge lean, their model operates at the word level. Since it does not take syntax into account, it has no means of deleting constituents spanning several subtrees (*e.g.,* relative clauses). Clarke and Lapata (2008) show that such unsupervised models can be greatly improved when linguistically motivated constraints are used during decoding.

## 3. Problem Formulation

As mentioned earlier, we formulate sentence compression as a tree-to-tree rewriting problem using a weighted synchronous grammar coupled with a large margin training process. Our model learns from a parallel corpus of input (uncompressed) and output (compressed) pairs $(\mathbf{x}_1, \mathbf{y}_1), \ldots, (\mathbf{x}_n, \mathbf{y}_n)$ to predict a target labeled tree $\mathbf{y}$ from a source labeled tree $\mathbf{x}$. We capture the dependency between $\mathbf{x}$ and $\mathbf{y}$ as a weighted STSG which we define in the following section. Section 3.2 discusses how we extract such a grammar from a parallel corpus. Each rule has a score, as does each ngram in the output tree, from which the overall score of a compression $\mathbf{y}$ for sentence $\mathbf{x}$ can be derived. We introduce our scoring function in Section 3.3 and explain our training algorithm in Section 3.5. In this framework decoding amounts to finding the best target tree licensed by the grammar given a source tree. We present a chart-based decoding algorithm in Section 3.4.

### 3.1 Synchronous Grammar

A synchronous grammar defines a space of valid source and target tree pairs, much as a regular grammar defines a space of valid trees. Synchronous grammars can be treated as tree transducers by reasoning over the space of possible sister trees for a given tree, that is, all the trees which can be produced alongside the given tree. This is essentially a transducer

---

**Algorithm 1** Generative process for creating a *pair of trees*.

---
initialize source tree, $\mathbf{x} = R_S$
initialize target tree, $\mathbf{y} = R_T$
initialize stack of frontier nodes, $F = [(R_S, R_T)]$
**for all** node pairs, $(v_S, v_T) \in F$ **do**
    choose a rule $\langle v_S, v_T \rangle \to \langle \alpha, \gamma, \sim \rangle$
    rewrite node $v_S$ in $\mathbf{x}$ as $\alpha$
    rewrite node $v_T$ in $\mathbf{y}$ as $\gamma$
    **for all** variables, $u \in \sim$ **do**
        find aligned child nodes, $(c_S, c_T)$, under $v_S$ and $v_T$ corresponding to $u$
        push $(c_S, c_T)$ on to $F$
    **end for**
**end for**
$\mathbf{x}$ and $\mathbf{y}$ are now complete

---

which takes a tree as input and produces a tree as output. The grammar rules specify the steps taken by the transducer in recursively mapping tree fragments of the input tree into fragments in the target tree. From the many families of synchronous grammars (see Section 2), we elect to use a synchronous tree-substitution grammar (STSG). This is one of the simpler formalisms, and consequently has efficient inference algorithms, while still being complex enough to model a rich suite of tree edit operations.

A STSG is a 7-tuple, $G = (\mathcal{N}_S, \mathcal{N}_T, \Omega_S, \Omega_T, P, R_S, R_T)$ where $\mathcal{N}$ are the non-terminals and $\Omega$ are the terminals, with the subscripts $S$ and $T$ indicating source and target respectively, $P$ are the productions and $R_S \in \mathcal{N}_S$ and $R_T \in \mathcal{N}_T$ are the distinguished root symbols. Each production is a rewrite rule for two aligned non-terminals $X \in \mathcal{N}_S$ and $Y \in \mathcal{N}_T$ in the source and target:

$$\langle X, Y \rangle \to \langle \alpha, \gamma, \sim \rangle \tag{1}$$

where $\alpha$ and $\gamma$ are *elementary trees* rooted with the symbols $X$ and $Y$ respectively. Note that a synchronous context free grammar (SCFG) limits $\alpha$ and $\gamma$ to one level elementary trees, but is otherwise identical to a STSG, which imposes no such limits. Non-terminal leaves of the elementary trees are referred to as *frontier nodes* or *variables*. These are the points of recursion in the transductive process. A one-to-one alignment between the frontier nodes in $\alpha$ and $\gamma$ is specified by $\sim$. The alignment can represent deletion (or insertion) by aligning a node with the special $\epsilon$ symbol, which indicates that the node is not present in the other tree. Only nodes in $\alpha$ can be aligned to $\epsilon$, which allows for subtrees to be deleted during transduction. We disallow the converse, $\epsilon$-aligned nodes in $\gamma$, as these would license unlimited insertion in the target tree, independently of the source tree. This capability would be of limited use for sentence compression, while also increasing the complexity of inference.

The grammar productions can be used in a generative setting to produce pairs of trees, or in a transductive setting to produce a target tree when given a source tree. Algorithms 1 and 2 present pseudo-code for both processes. The generative process (Algorithm 1) starts with the two root symbols and applies a production which rewrites the symbols as the production's elementary trees. These elementary trees might contain frontier nodes, in

---

**Algorithm 2** The *transduction* of a source tree into a target tree.

---

**Require:** complete source tree, **x**, with root node labeled $R_S$

 initialize target tree, $\mathbf{y} = R_T$

 initialize stack of frontier nodes, $F = [(root(\mathbf{x}), R_T)]$

 **for all** node pairs, $(v_S, v_T) \in F$ **do**

  choose a rule $\langle v_S, v_T \rangle \to \langle \alpha, \gamma, \sim \rangle$ where $\alpha$ matches the sub-tree rooted at $v_S$ in **x**

  rewrite $v_T$ as $\gamma$ in **y**

  **for all** variables, $u \in \sim$ **do**

   find aligned child nodes, $(c_S, c_T)$, under $v_S$ and $v_T$ corresponding to $u$

   push $(c_S, c_T)$ on to $F$

  **end for**

 **end for**

 **y** is now complete

---

which case the aligned pairs of frontier nodes are pushed on to the stack, and later rewritten using another production. The process continues in a recursive fashion until the stack is empty — there are no frontier nodes remaining —, at which point the two trees are complete. The sequence of rewrite rules are referred to as a *derivation*, from which the source and target tree can be recovered deterministically.

Our model uses a STSG in a transductive setting, where the source tree is given and it is only the target tree that is generated. This necessitates a different rewriting process, as shown in Algorithm 2. We start with the source tree, and $R_T$, the target root symbol, which is aligned to the root node of the source, denoted $root(\mathbf{x})$. Then we choose a production to rewrite the pair of aligned non-terminals such that the production's source side, $\alpha$, matches the source tree. The target symbol is then rewritten using $\gamma$. For each variable in $\alpha$ the matching node in the source and its corresponding leaf node in the target tree are pushed on to the stack for later processing.[1] The process repeats until the stack is empty, and therefore the source tree has been covered. We now have a complete target tree. As before we use the term *derivation* to refer to this sequence of production applications. The target string is the *yield* of the target tree, given by reading the non-terminals from the tree in a left to right manner.

Let us consider again the compression example from Figure 1. The tree editing rules from Figure 2 are encoded as STSG productions in Figure 3 (see rules (1)–(5)). Production (1), reproduces tree pair (1) from Figure 2, production (2) tree pair (2), and so on. The notation in Figure 3 (primarily for space reasons) uses brackets ([]) to indicate constituent boundaries. Brackets surround a constituent's non-terminal and its child nodes, which can each be terminals, non-terminals or bracketed subtrees. The boxed indices are short-hand notation for the alignment, $\sim$. For example, in rule (1) they specify that the two WP non-terminals are aligned and the RB node occurs only in the source tree (i.e., heads a deleted sub-tree). The grammar rules allow for differences in non-terminal category between the source and target, as seen in rules (2)–(4). They also allow arbitrarily deep elementary trees, as evidenced by rule (5) which is has trees of depth two. Rules (6)–(12) complete the toy grammar which describes the tree pair from Figure 1. These rules copy parts of the source tree into the target, be they terminals (e.g., rule (6)) or internal nodes with children (*e.g.,* rule (9)).

---

1. Special care must be taken for $\epsilon$ aligned variables. Nodes in $\alpha$ which are $\epsilon$-aligned signify that the source sub-tree below this point can be deleted without affecting the target tree. For this reason we can safely ignore source nodes deleted in this manner.

| | | | |
|---|---|---|---|
| | *Rules which perform major tree edits* | | |
| (1) | $\langle$WHNP, WHNP$\rangle$ | $\rightarrow$ | $\langle$[WHNP RB$_4$ WP$_1$], [WHNP WP$_1$]$\rangle$ |
| (2) | $\langle$S, NP$\rangle$ | $\rightarrow$ | $\langle$[S NP$_1$ VP$_4$], NP$_1$]$\rangle$ |
| (3) | $\langle$S, VP$\rangle$ | $\rightarrow$ | $\langle$[S NP$_4$ VP$_1$], VP$_1$]$\rangle$ |
| (4) | $\langle$S̄, VP$\rangle$ | $\rightarrow$ | $\langle$[S̄ WHNP$_4$ S$_1$], VP$_1$]$\rangle$ |
| (5) | $\langle$S, S$\rangle$ | $\rightarrow$ | $\langle$[S [S̄ WHNP$_1$ S$_2$] [CC and] S̄$_3$], [S WHNP$_1$ [S NP$_2$ VP$_3$]]$\rangle$ |
| | *Rules which preserve the tree structure* | | |
| (6) | $\langle$WP, WP$\rangle$ | $\rightarrow$ | $\langle$[WP what], [WP what]$\rangle$ |
| (7) | $\langle$NP, NP$\rangle$ | $\rightarrow$ | $\langle$[NP NNS$_1$], [NP NNS$_1$]$\rangle$ |
| (8) | $\langle$NNS, NNS$\rangle$ | $\rightarrow$ | $\langle$[NNS records], [NNS records]$\rangle$ |
| (9) | $\langle$VP, VP$\rangle$ | $\rightarrow$ | $\langle$[VP VBP$_1$ VP$_2$], [VP VBP$_1$ VP$_2$]$\rangle$ |
| (10) | $\langle$VBP, VBP$\rangle$ | $\rightarrow$ | $\langle$[VBP are], [VBP are]$\rangle$ |
| (11) | $\langle$VP, VP$\rangle$ | $\rightarrow$ | $\langle$[VP VBN$_1$], [VP VBN$_1$]$\rangle$ |
| (12) | $\langle$VBN, VBN$\rangle$ | $\rightarrow$ | $\langle$[VBN involved], [VBN involved]$\rangle$ |

Figure 3: The rules in a Synchronous Tree Substitution Grammar (STSG) capable of generating the sentence pair from Figure 1. Equivalently, this grammar defines a transducer which can convert the source tree (Figure 1(a)) into the target tree (Figure 1(b)). Each rule rewrites a pair of non-terminals into a pair of subtrees, shown in bracketed notation.

Figure 4 shows how this grammar can be used to transduce the source tree into the target tree from Figure 1. The first few steps of the derivation are also shown graphically in Figure 5. We start with the source tree, and seek to transduce its root symbol into the target root symbol, denoted S/S. The first rule to be applied is rule (5) in Figure 3; its source side, $\alpha = $ [S [S̄ WHNP S] [CC and] S̄], matches the root of source tree and it has the requisite target category, $Y = S$. The matching part of the source tree is rewritten using the rule's target elementary tree, $\gamma = $ [S WHNP [S NP VP]]. The three three variables are now annotated to reflect the category transformations required for each node, WHNP/WHNP, S/NP and S̄/VP. The process now continues for the leftmost of these nodes, labeled WHNP/WHNP. Rule (1) (from Figure 3) is then applied, which deletes the node's left child, shown as RB/$\epsilon$, and retains its right child. The subsequent rule completes the transduction of the WHNP node by matching the string '*exactly*'. The algorithm continues to visit each variable node and finishes when there are no variable nodes remaining, resulting in the desired target tree.

## 3.2 Grammar

The previous section outlined the STSG formalism we employ in our sentence compression model, save one important detail: the grammar itself. For example, we could obtain a

```
[S/S [S̄ [WHNP exactly what] [S [NP records] [VP made it]]]
       [CC and] [S̄ [WHNP which] [S [NP ones] [VP are involved]]]]]
⇒₅   [S [WHNP/WHNP [RB exactly] [WP what]] [S [S/NP [NP records] [VP made it]]
       [S̄/VP [WHNP which] [S [NP ones] [VP are involved]]]]]]
⇒₁   [S [WHNP [WP/WP what]] [S [S/NP [NP records] [VP made it]]
       [S̄/VP [WHNP which] [S [NP ones] [VP are involved]]]]]]
⇒₆   [S [WHNP [WP what]] [S [S/NP [NP records] [VP [VBD made] [NP [PRP it]]]]]
       [S̄/VP [WHNP which] [S [NP ones] [VP [VBP are] [VP [VBN involved]]]]]]]]
⇒₂   [S [WHNP [WP what]] [S [NP [NNS/NNS records]]
       [S̄/VP [WHNP which] [S [NP ones] [VP are involved]]]]]]
⇒₈   [S [WHNP [WP what]] [S [NP [NNS records]]]
       [S̄/VP [WHNP which] [S [NP ones] [VP are involved]]]]]
⇒₄   [S [WHNP what] [S [NP records] [S/VP [NP ones] [VP are involved]]]]]
⇒₃   [S [WHNP what] [S [NP records] [VP/VP [VP [VBP are] [VP [VBN involved]]]]]]]
⇒₉   [S [WHNP what] [S [NP records] [VP [VBP/VBP are] [VP/VP [VBN involved]]]]]]
⇒₁₀  [S [WHNP what] [S [NP records] [VP [VBP are] [VP/VP [VBN involved]]]]]
⇒₁₁  [S [WHNP what] [S [NP records] [VP [VBP are] [VP [VBN/VBN involved]]]]]]
⇒₁₂  [S [WHNP [WP what]] [S [NP [NNS records]] [VP [VBP are] [VP [VBN involved]]]]]]
```

Figure 4: Derivation of example sentence pair from Figure 1. Each line shows a rewrite step, denoted $\Rightarrow_i$ where the subscript $i$ identifies which rule was used. The frontier nodes are shown in bold with $X/Y$ indicating that symbol $X$ must be transduced into $Y$ in subsequent steps. For the sake of clarity, some internal nodes have been omitted.

synchronous grammar by hand, automatically from a corpus, or by some combination. Our only requirement is that the grammar allows the source trees in the training set to be transduced into their corresponding target trees. For maximum generality, we devised an automatic method to extract a grammar from a parsed, word-aligned parallel compression corpus. The method maps the word alignment into a constituent level alignment between nodes in the source and target trees. Pairs of aligned subtrees are next generalized to create tree fragments (elementary trees) which form the rules of the grammar.

The first step of the algorithm is to find the constituent alignment, which we define as the set of source and target constituent pairs whose yields are aligned to one another under the word alignment. We base our approach on the alignment template method (Och & Ney, 2004), which uses word alignments to define alignments between ngrams (called *phrases* in the SMT literature). This method finds pairs of ngrams where at least one word in one of the ngrams is aligned to a word in the other, but no word in either ngram is aligned to a word outside the other ngram. In addition, we require that these ngrams are syntactic constituents. More formally, we define constituent alignment as:

$$\mathcal{C} = \{(v_S, v_T), \ (\exists(s,t) \in \mathcal{A} \land s \in Y(v_S) \land t \in Y(v_T)) \land \tag{2}$$
$$(\nexists(s,t) \in \mathcal{A} \land (s \in Y(v_S) \veebar t \in Y(v_T)))\}$$

where $v_S$ and $v_T$ are source and target tree nodes (subtrees), $\mathcal{A} = \{(s,t)\}$ is the set of word alignments (pairs of word-indices), $Y(\cdot)$ returns the yield span for a subtree (the minimum and maximum word index in its yield) and $\veebar$ is the exclusive-or operator. Figure 6 shows the word alignment and the constituent alignments that are licensed for the sentence pair from Figure 1.

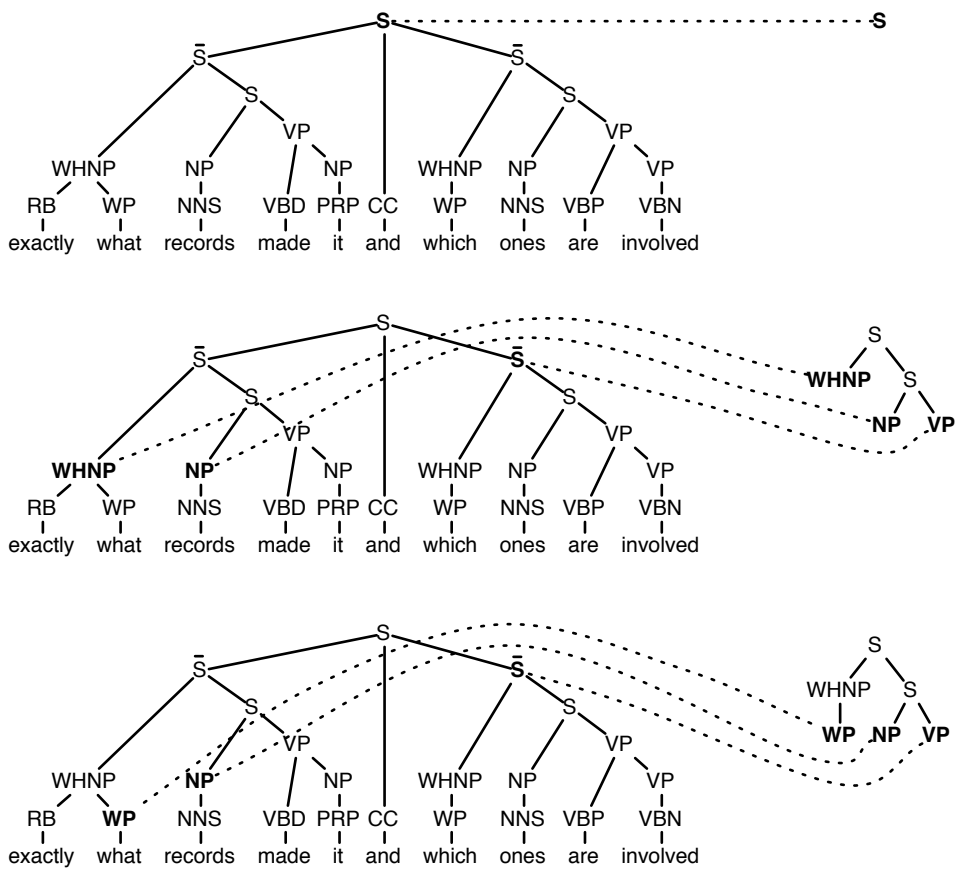

Figure 5: Graphical depiction of the first two steps of the derivation in Figure 4. The source tree is shown on the left and the partial target tree on the right. Variable nodes are shown in bold face and dotted lines show their alignment.

The next step is to generalize the aligned subtree pairs by replacing aligned child subtrees with variable nodes. For example, in Figure 6 when we consider the pair of aligned subtrees [S̄ which ones are involved] and [VP are involved], we could extract the rule:

$$\langle \bar{S}, VP \rangle \rightarrow \langle [\bar{S} \; [WHNP \; [WP \; which]] \; [S \; [NP \; [NNS \; ones]] \; [VP \; [VBP \; are] \; [VP \; [VBN \; involved]]]]]], \\ [VP \; [VBP \; are] \; [VP \; [VBN \; involved]]]] \rangle \qquad (3)$$

However, this rule is very specific and consequently will not be very useful in a transduction model. In order for it to be applied, we must see the full S̄ subtree, which is highly unlikely to occur in another sentence. Ideally, we should generalize the rule so as to match many more source trees, and thereby allow transduction of previously unseen structures. In the example, the node pairs labeled (VP$_1$, VP$_1$), (VBP, VBP), (VP$_2$, VP$_2$) and (VBN, VBN) can all be generalized as these nodes are aligned constituents (subscripts added to distinguish between the two VP nodes). In addition, the nodes WHNP, WP, NP and NNS in the source are unaligned, and therefore can be generalized using $\epsilon$-alignment to signify deletion. If we were to perform all possible generalizations for the above example,[2] we would produce the rule:

$$\langle \bar{S}, VP \rangle \rightarrow \langle [\bar{S} \ WHNP_{\boxed{A}} \ S_{\boxed{1}}], VP_{\boxed{1}} \rangle \tag{4}$$

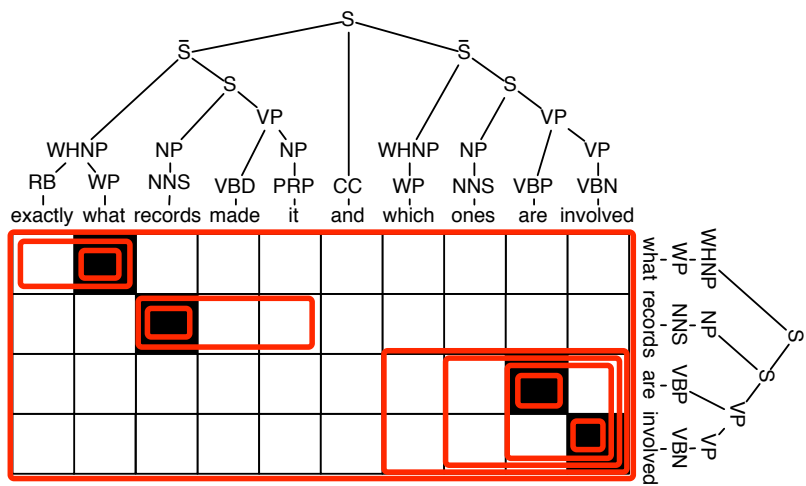

Figure 6: Tree pair with word alignments shown as a binary matrix. A dark square indicates an alignment between the words on its row and column. The overlaid rectangles show constituent alignments which are inferred from the word alignment.

There are many other possible rules which can be extracted by applying different legal combinations of the generalizations (there are 45 in total for this example).

Algorithm 3 shows how the minimial (most general) rules are extracted.[3] This results in the *minimal set* of synchronous rules which can describe each tree pair.[4] These rules are minimal in the sense that they cannot be made smaller (e.g., by replacing a subtree with a variable) while still honoring the word-alignment. Figure 7 shows the resulting minimal set of synchronous rules for the example from Figure 6. As can be seen from the example, many of the rules extracted are overly general. Ideally, we would extract every rule with every legal combination of generalizations, however this leads to a massive number of rules — exponential in the size of the source tree. We address this problem by allowing a limited number of generalizations to be 'skipped' in the extraction process. This is equivalent to altering lines 4 and 7 in Algorithm 3 to first make a non-deterministic decision whether to match or ignore the match and continue descending the source tree. The recursion depth limits the number of matches that can be ignored in this way. For example, if we allow one

---

2. Where some generalizations are mutually exclusive, we take the highest match in the trees.
3. The non-deterministic matching step in line 8 allows the matching of *all* options individually. This is implemented as a mutually recursive function which replicates the algorithm state to process each different match.
4. Algorithm 3 is an extension of Galley, Hopkins, Knight, and Marcu's (2004) technique for extracting a SCFG from a word-aligned corpus consisting of (tree, string) pairs.

---

**Algorithm 3** $extract(\mathbf{x}, \mathbf{y}, A)$: extracts minimal rules from constituent-aligned trees

---

**Require:** source tree, $\mathbf{x}$, target tree, $\mathbf{y}$, and constituent-alignment, $A$

 1: initialize source and target sides of rule, $\alpha = \mathbf{x}, \gamma = \mathbf{y}$

 2: initialize frontier alignment, $\sim = \emptyset$

 3: **for all** nodes $v_S \in \alpha$, top-down **do**

 4:    **if** $v_S$ is null-aligned **then**

 5:       $\sim \leftarrow \sim \cup (v_S, \epsilon)$

 6:       delete children of $a$

 7:    **else if** $v_S$ is aligned to some target node(s) **then**

 8:       choose target node, $v_T$                     {non-deterministic choice}

 9:       call $extract(v_S, v_T, A)$

10:       $\sim \leftarrow \sim \cup (v_S, v_T)$

11:       delete children of $v_S$

12:       delete children of $v_T$

13:    **end if**

14: **end for**

15: emit rule $\langle root(\alpha), root(\gamma) \rangle \rightarrow \langle \alpha, \gamma, \sim \rangle$

---

level of recursion when extracting rules from the $(\bar{\mathrm{S}}, \mathrm{VP})$ pair from Figure 6, we get the additional rules:

$$\langle \bar{\mathrm{S}}, \mathrm{VP} \rangle \rightarrow \langle [\bar{\mathrm{S}}\ [\mathrm{WHNP}\ \mathrm{WP}_{\boxed{6}}\ \mathrm{S}_{\boxed{1}}],\ \mathrm{VP}_{\boxed{1}} \rangle$$

$$\langle \bar{\mathrm{S}}, \mathrm{VP} \rangle \rightarrow \langle [\bar{\mathrm{S}}\ \mathrm{WHNP}_{\boxed{6}}\ [\mathrm{S}\ \mathrm{NP}_{\boxed{6}} \mathrm{VP}_{\boxed{1}}]],\ \mathrm{VP}_{\boxed{1}} \rangle$$

while at two levels of recursion, we also get:

$$\langle \bar{\mathrm{S}}, \mathrm{VP} \rangle \rightarrow \langle [\bar{\mathrm{S}}\ [\mathrm{WHNP}\ [\mathrm{WP}\ \mathrm{which}]]\ \mathrm{S}_{\boxed{1}}],\ \mathrm{VP}_{\boxed{1}} \rangle$$

$$\langle \bar{\mathrm{S}}, \mathrm{VP} \rangle \rightarrow \langle [\bar{\mathrm{S}}\ [\mathrm{WHNP}\ [\mathrm{WP}\ \mathrm{which}]]\ [\mathrm{S}\ \mathrm{NP}_{\boxed{6}} \mathrm{VP}_{\boxed{1}}]],\ \mathrm{VP}_{\boxed{1}} \rangle$$

$$\langle \bar{\mathrm{S}}, \mathrm{VP} \rangle \rightarrow \langle [\bar{\mathrm{S}}\ \mathrm{WHNP}_{\boxed{6}}\ [\mathrm{S}\ [\mathrm{NP}\ \mathrm{NNS}_{\boxed{6}}]\ \mathrm{VP}_{\boxed{1}}]],\ \mathrm{VP}_{\boxed{1}} \rangle$$

$$\langle \bar{\mathrm{S}}, \mathrm{VP} \rangle \rightarrow \langle [\bar{\mathrm{S}}\ \mathrm{WHNP}_{\boxed{6}}\ [\mathrm{S}\ \mathrm{NP}_{\boxed{6}} [\mathrm{VP}\ \mathrm{VBD}_{\boxed{1}}\ \mathrm{VP}_{\boxed{2}}]]],\ [\mathrm{VBD}_{\boxed{1}}\ \mathrm{VBD}_{\boxed{2}}] \rangle$$

Compared to rule (4) we can see that the specialized rules above add useful structure and lexicalisation, but are still sufficiently abstract to generalize to new sentences, unlike rule (3). The number of rules is exponential in the recursion depth, but with fixed a depth it is polynomial in the size of the source tree fragment. We set the recursion depth to a small number (one or two) in our experiments.

There is no guarantee that the induced rules will have good coverage on unseen trees. Tree fragments containing previously unseen terminals or non-terminals, or even an unseen sequence of children for a parent non-terminal, cannot be matched by any grammar productions. In this case the transduction algorithm (Algorithm 2) will fail as it has no way of covering the source tree. However, the problem can be easily remedied by adding new rules to the grammar to allow the source tree to be fully covered.[5] For each node in the

---

5. There are alternative, equally valid, techniques for improving coverage which simplify the syntax trees. For example, this can be done explicitly by binarizing large productions (e.g., Petrov, Barrett, Thibaux, & Klein, 2006) or implicitly with a Markov grammar over grammar productions (e.g., Collins, 1999).

$$\langle S,S \rangle \rightarrow \langle [S\ [\bar{S}\ WHNP_{\boxed{1}}\ S_{\boxed{2}}]\ CC_{\boxed{c}}\ \bar{S}_{\boxed{3}}],\ [S\ WHNP_{\boxed{1}}\ [S\ NP_{\boxed{2}}\ VP_{\boxed{3}}]] \rangle$$
$$\langle WHNP,WHNP \rangle \rightarrow \langle [WHNP\ RB_{\boxed{c}}WP_{\boxed{1}}],\ [WHNP\ WP_{\boxed{1}}] \rangle$$
$$\langle WP,WP \rangle \rightarrow \langle [WP\ what],\ [WP\ what] \rangle$$
$$\langle S,NP \rangle \rightarrow \langle [S\ NP_{\boxed{1}}\ VP_{\boxed{c}}],\ NP_{\boxed{1}} \rangle$$
$$\langle NP,NP \rangle \rightarrow \langle [NP\ NNS_{\boxed{1}}],\ [NP\ NNS_{\boxed{1}}] \rangle$$
$$\langle NNS,NNS \rangle \rightarrow \langle [NNS\ records],\ [NNS\ records] \rangle$$
$$\langle \bar{S},VP \rangle \rightarrow \langle [\bar{S}\ WHNP_{\boxed{c}}\ S_{\boxed{1}}],\ VP_{\boxed{1}} \rangle$$
$$\langle S,VP \rangle \rightarrow \langle [S\ NP_{\boxed{c}}\ VP_{\boxed{1}}],\ VP_{\boxed{1}} \rangle$$
$$\langle VP,VP \rangle \rightarrow \langle [VP\ VBP_{\boxed{1}}\ VP_{\boxed{2}}],\ [VP\ VBP_{\boxed{1}}\ VP_{\boxed{2}}] \rangle$$
$$\langle VBP,VBP \rangle \rightarrow \langle [VBP\ are],\ [VBP\ are] \rangle$$
$$\langle VP,VP \rangle \rightarrow \langle [VP\ VBN_{\boxed{1}}],\ [VP\ VBN_{\boxed{1}}] \rangle$$
$$\langle VBN,VBN \rangle \rightarrow \langle [VBN\ involved],\ [VBN\ involved] \rangle$$

Figure 7: The minimal set of STSG rules extracted from the aligned trees in Figure 6.

source tree, a rule is created to copy that node and its child nodes into the target tree. For example, if we see the fragment [NP DT JJ NN] in the source tree, we add the rule:

$$\langle NP,NP \rangle \rightarrow \langle [NP\ DT_{\boxed{1}}\ JJ_{\boxed{2}}\ NN_{\boxed{3}}],\ [NP\ DT_{\boxed{1}}\ JJ_{\boxed{2}}\ NN_{\boxed{3}}] \rangle$$

With these rules, each source node is copied into the target tree, and therefore the transduction algorithm can trivially recreate the original tree. Of course, the other grammar rules can work in conjunction with the copying rules to produce other target trees.

While the copy rules solve the coverage problem on unseen data, they do not solve the related problem of under-compression. This occurs when there are unseen CFG productions in the source tree and therefore the only applicable grammar rules are copy rules, which copy all child nodes into the target. None of the child subtrees can be deleted unless the parent node can itself deleted by a higher-level rule, in which case all the children are deleted. Clearly, it would add considerable modelling flexibility to be able to delete some, but not all, of the children. For this reason, we add explicit deletion rules for each source CFG production which allow subsets of the child nodes to be deleted in a linguistically plausible manner.

The deletion rules attempt to preserve the most important child nodes. We measure importance using the head-finding heuristic from Collins' parser (Appendix A, Collins, 1999). Collins' method finds the single head child of a CFG production using hand-coded tables for each non-terminal type. As we desire a set of child nodes, we run the algorithm to find all matches rather than stopping after the first match. The order in which each match is found is used as a ranking of the importance of each child. The ordered list of child nodes is then used to create synchronous rules which retain head 1, heads 1–2, ..., all heads.

For the fragment [NP DT JJ NN], the heads are found in the following order (NN, DT, JJ). Therefore we create rules to retain children (NN); (DT, NN) and (DT, JJ, NN):

$$\langle \text{NP,NP} \rangle \rightarrow \langle [\text{NP DT}_{\boxed{4}} \text{ JJ}_{\boxed{4}} \text{ NN}_{\boxed{1}}], [\text{NP NN}_{\boxed{1}}] \rangle$$
$$\langle \text{NP,NN} \rangle \rightarrow \langle [\text{NP DT}_{\boxed{4}} \text{ JJ}_{\boxed{4}} \text{ NN}_{\boxed{1}}], \text{NN}_{\boxed{1}} \rangle$$
$$\langle \text{NP,NP} \rangle \rightarrow \langle [\text{NP DT}_{\boxed{1}} \text{ JJ}_{\boxed{4}} \text{ NN}_{\boxed{2}}], [\text{NP DT}_{\boxed{1}} \text{ NN}_{\boxed{2}}] \rangle$$
$$\langle \text{NP,NP} \rangle \rightarrow \langle [\text{NP DT}_{\boxed{1}} \text{ JJ}_{\boxed{2}} \text{ NN}_{\boxed{3}}], [\text{NP DT}_{\boxed{1}} \text{ JJ}_{\boxed{2}} \text{ NN}_{\boxed{3}}] \rangle$$

Note that when only one child remains, the rule is also produced without the parent node, as seen in the second rule above.

### 3.3 Linear Model

While an STSG defines a transducer capable of mapping a source tree into many possible target trees, it is of little use without some kind of weighting towards grammatical trees which have been constructed using sensible STSG productions and which yield fluent compressed target sentences. Ideally the model would define a scoring function over target trees or strings, however we instead operate on derivations. In general, there may be many derivations which all produce the same target tree, a situation referred to as *spurious ambiguity*. To fully account for spurious ambiguity would require aggregating all derivations which produce the same target tree. This would break the polynomial-time dynamic program used for inference, rendering inference problem NP-complete (Knight, 1999). To this end, we define a scoring function over derivations:

$$score(\mathbf{d}; \mathbf{w}) = \langle \Psi(\mathbf{d}), \mathbf{w} \rangle \tag{5}$$

where $\mathbf{d}$ is a derivation[6] consisting of a sequence of rules, $\mathbf{w}$ are the model parameters, $\Psi$ is a vector-valued feature function and the operator $\langle \cdot, \cdot \rangle$ is the inner product. The parameters, $\mathbf{w}$, are learned during training, described in Section 3.5.

The feature function, $\Psi$, is defined as:

$$\Psi(\mathbf{d}) = \sum_{r \in \mathbf{d}} \phi(r, source(\mathbf{d})) + \sum_{m \in ngrams(\mathbf{d})} \psi(m, source(\mathbf{d})) \tag{6}$$

where $r$ are the rules of a derivation, $ngrams(\mathbf{d})$ are the ngrams in the yield of the target tree and $\phi$ is a feature function returning a vector of feature values for each rule. Note that the feature function has access to not only the rule, $r$, but also the source tree, $source(\mathbf{d})$, as this is a conditional model and therefore doing so has no overhead in terms of modeling assumptions or the complexity of inference.

In the second summand in (6), $m$ are the ngrams in the yield of the target tree and $\psi$ is a feature function over these ngrams. Traditional (weighted) synchronous grammars only allow features which decompose with the derivation (i.e., can be expressed using the first summand in (6)). However, this is a very limiting requirement, as the ngram features allow the modeling of local coherence and are commonly used in the sentence compression literature (Knight & Marcu, 2002; Turner & Charniak, 2005; Galley & McKeown, 2007;

---

6. The derivation, $\mathbf{d}$, fully specifies both the source, $\mathbf{x} = source(\mathbf{d})$, and the target tree, $\mathbf{y} = target(\mathbf{d})$.

Clarke & Lapata, 2008; Hori & Furui, 2004; McDonald, 2006). For instance, when deleting a sub-tree with left and right siblings, it is critical to know not only that the new siblings are in a grammatical configuration, but also that their yield still forms a coherent string. For this reason, we allow ngram features, specifically the conditional log-probability of an ngram language model. Unfortunately, this comes at a price as the ngram features significantly increase the complexity of inference used for training and decoding.

### 3.4 Decoding

Decoding aims to find the best target tree licensed by the grammar given a source tree. As mentioned above, we deal with derivations in place of target trees. Decoding finds the maximizing derivation, $\mathbf{d}^*$, of:

$$\mathbf{d}^* = \operatorname*{argmax}_{\mathbf{d}:source(\mathbf{d})=\mathbf{x}} score(\mathbf{d}; \mathbf{w}) \qquad (7)$$

where $\mathbf{x}$ is the (given) source tree, $source(\mathbf{d})$ extracts the source tree from the derivation $\mathbf{d}$ and $score$ is defined in (5). The maximization is performed over the space of derivations for the given source tree, as defined by the transduction process shown in Algorithm 2.

The maximization problem in (7) is solved using the chart-based dynamic program shown in Algorithm 4. This extends earlier inference algorithms for weighted STSGs (Eisner, 2003) which assume that the scoring function must decompose with the derivation, i.e., features apply to rules but not to terminal ngrams. Relaxing this assumption leads to additional complications and increased time and space complexity. This is equivalent to using as our grammar the intersection between the original grammar and an ngram language model, as explained by Chiang (2007) in the context of string transduction with an SCFG.

The algorithm defines a chart, $C$, to record the best scoring (partial) target tree for each source node $v_S$ and with root non-terminal $t$. The back-pointers, $B$, record the maximizing rule and store pointers to the child chart cells filling each variable in the rule. The chart is also indexed by the $n - 1$ terminals at the left and right edges of the target tree's yield to allow scoring of ngram features.[7] The terminal ngrams provide sufficient context to evaluate ngram features overlapping the cell's boundary when the chart cell is combined in another rule application (this is the operation performed by the boundary-ngrams function on line 15). This is best illustrated with an example. Using trigram features, $n = 3$, if a node were rewritten as [NP the fast car] then we must store the ngram context (the fast, fast car) in its chart entry. Similarly [VP skidded to a halt] would have ngram context (skidded to, a halt). When applying a parent rule [S NP VP] which rewrites these two trees as adjacent siblings we need to find the ngrams on the boundary between the NP and VP. These can be easily retrieved from the two chart cells' contexts. We combine the right edge of the NP context, 'fast car', with the left edge of the VP context, 'skidded to', to get the two trigrams 'fast car skidded' and 'car skidded to'. The other trigrams — 'the fast car', 'skidded to a' and 'to a halt' — will have already been evaluated in the child chart cells. The new combined S chart cell is now given the context (the fast, a halt) by taking the left and right

---

7. Strictly speaking, only the terminals on the right edge are required for a compression model which would create the target string in a left-to-right manner. However, our algorithm is more general in that it allows reordering rules such as $\langle \text{PP,PP} \rangle \rightarrow \langle [\text{PP IN}_{\boxed{1}} \text{ NP}_{\boxed{2}}], [\text{PP NP}_{\boxed{2}} \text{ IN}_{\boxed{1}}] \rangle$. Such rules are required for most other text-rewriting tasks besides sentence compression.

---

**Algorithm 4** Exact chart based decoding algorithm.

---

**Require:** complete source tree, $\mathbf{x}$, with root node labeled $R_S$

 1: let $C[v, t, l] \in \mathcal{R}$ be a chart representing the score of the best derivation transducing the tree rooted at $v$ to a tree with root category $t$ and ngram context $l$

 2: let $B[v, t, l] \in (P, \mathbf{x} \times \mathcal{N}_T \times \mathcal{L})$ be the corresponding back-pointers, each consisting of a production and the source node, target category and ngram context for each of the production's variables

 3: initialize chart, $C[*, *, *] = -\infty$

 4: initialize back-pointers, $B[*, *, *] = $ none

 5: **for all** source nodes, $v_S \in \mathbf{x}$, bottom-up **do**

 6:     **for all** rules, $r = \langle v_S, Y \rangle \rightarrow \langle \alpha, \gamma, \sim \rangle$ where $\alpha$ matches the sub-tree rooted at $v_S$ **do**

 7:         let $m$ be the target ngrams wholly contained in $\gamma$

 8:         let features vector, $\Psi \leftarrow \phi(r, \mathbf{x}) + \psi(m, \mathbf{x})$

 9:         let $l$ be an empty ngram context

10:         let score, $q \leftarrow 0$

11:         **for all** variables, $u \in \sim$ **do**

12:             find source child node, $c_u$, under $v_S$ corresponding to $u$

13:             let $t_u$ be the non-terminal for target child node under $\gamma$ corresponding to $u$

14:             choose child chart entry, $q_u = C[c_u, t_u, l_u]$         {non-deterministic choice of $l_u$}

15:             let $m \leftarrow$ boundary-ngrams$(r, l_u)$

16:             update features, $\Psi \leftarrow \Psi + \psi(m, \mathbf{x})$

17:             update ngram context, $l \leftarrow$ merge-ngram-context$(l, l_u)$

18:             update score, $q \leftarrow q + q_u$

19:         **end for**

20:         update score, $q \leftarrow q + \langle \Psi, \mathbf{w} \rangle$

21:         **if** $q > C[v_S, Y, l]$ **then**

22:             update chart, $C[v_S, Y, l] \leftarrow q$

23:             update back-pointers, $B[v_S, Y, l] \leftarrow (r, \{(c_u, t_u, l_u) \forall u\})$

24:         **end if**

25:     **end for**

26: **end for**

27: find best root chart entry, $l^* \leftarrow \text{argmax}_l C[root(\mathbf{x}), R_T, l]$

28: create derivation, $\mathbf{d}$, by traversing back-pointers from $B[root(\mathbf{x}), R_T, l^*]$

---

edges of the two child cells. This merging process is performed by the merge-ngram-context function on line 17. Finally we add artificial root node to the target tree with $n-1$ artificial start terminals and one end terminal. This allows the ngram features to be applied over boundary ngrams at the beginning and end of the target string.

The decoding algorithm processes the source tree in a post-order traversal, finding the set of possible trees and their ngram contexts for each source node and inserting these into the chart. The rules which match the node are processed in lines 6–24. The feature vector, $\Psi$, is calculated on the rule and the ngrams therein (line 8), and for ngrams bordering child cells filling the rule's variables (line 16). Note that the feature vector only includes those features specific to the rule and the boundary ngrams, but not those wholly contained in the child cell. For this reason the score is the sum of the scores for each child cell (line 18) and the feature vector and the model weights (line 20). The new ngram context, $l$, is calculated by combining the rule's frontier and the ngram contexts of the child cells (line 17). Finally the chart entry for this node is updated if the score betters the previous value (lines 21–24).

When choosing the child chart cell entry in line 14, there can be many different entries each with a different ngram context, $l_u$. This affects the ngram features, $\psi$, and consequently the ngram context, $l$, and the score, $q$, for the rule. The non-determinism means that every combination of child chart entries are chosen for each variable, and these combinations are then evaluated and inserted into the chart. The number of combinations is the product of the number of child chart entries for each variable. This can be bounded by $O(|\mathcal{T}_T|^{2(n-1)V})$ where $|\mathcal{T}_T|$ is the size of the target lexicon and $V$ is the number of variables. Therefore the asymptotic time complexity of decoding is the $O(SR|\mathcal{T}_T|^{2(n-1)V})$ where $S$ are the number of source nodes and $R$ is the number of matching rules for each node. This high complexity clearly makes exact decoding infeasible, especially so when either $n$ or $V$ are large.

We adopt a popular approach in syntax-inspired machine translation to address this problem (Chiang, 2007). Firstly, we use a beam-search, which limits the number of different ngram contexts stored in each chart cell to a constant, $W$. This changes the base in the complexity term, leading to an improved $O(SRW^V)$ but which is still exponential in the number of variables. In addition, we use Chiang's cube-pruning heuristic to further limit the number of combinations. Cube-pruning uses a heuristic scoring function which approximates the conditional log-probability from a ngram language model with the log-probability from a unigram model.[8] This allows us to visit the combinations in best-first order under the heuristic scoring function until the beam is filled.The beam is then rescored using the correct scoring function. This can be done cheaply in $O(WV)$ time, leading to an overall time complexity of decoding to $O(SRWV)$. We refer the interested reader to the work of Chiang (2007) for further details.

### 3.5 Training

We now turn to the problem of how derivations are scored in our model. For a given source tree, the space of sister target trees implied by the synchronous grammar is often very large, and the majority of these trees are ungrammatical or poor compressions. It is the job of the training algorithm to find weights such that the reference target trees have high scores and the many other target trees licensed by the grammar are given lower scores.

As explained in Section 3.3 we define a scoring function over derivations. This function was given in (5) and (7), and is reproduced below:

$$f(\mathbf{d}; \mathbf{w}) = \underset{\mathbf{d}:source(\mathbf{d})=\mathbf{x}}{\operatorname{argmax}} \ \langle \mathbf{w}, \Psi(\mathbf{d}) \rangle \tag{8}$$

Equation (8) finds the best scoring derivation, $\mathbf{d}$, for a given source, $\mathbf{x}$, under a linear model. Recall that $\mathbf{y}$ is a derivation which generates the source tree $\mathbf{x}$ and a target tree. The goal

---

8. We use the conditional log-probability of an ngram language model as our only ngram feature. In order to use other ngram features, such as binary identity features for specific ngrams, it would first be advisable to construct an approximation which decomposes with the derivation for use in the cube-pruning heuristic.

of the training procedure is to find a parameter vector $\mathbf{w}$ which satisfies the condition:

$$\forall i, \forall \mathbf{d} : source(\mathbf{d}) = \mathbf{x}_i \land \mathbf{d} \neq \mathbf{d}_i : \langle \mathbf{w}, \Psi(\mathbf{d}_i) - \Psi(\mathbf{d}) \rangle \geq 0 \tag{9}$$

where $\mathbf{x}_i, \mathbf{d}_i$ are the $i$th training source tree and reference derivation. This condition states that for all training instances the reference derivation is at least as high scoring as any other derivations. Ideally, we would also like to know the extent to which a predicted target tree differs from the reference tree. For example, a compression that differs from the gold standard with respect to one or two words should be treated differently from a compression that bears no resemblance to it. Another important factor is the length of the compression. Compressions whose length is similar to the gold standard should be be preferable to longer or shorter output. A loss function $\Delta(\mathbf{y}_i, \mathbf{y})$ quantifies the accuracy of prediction $\mathbf{y}$ with respect to the true output value $\mathbf{y}_i$.

There are a plethora of different discriminative training frameworks which can optimize a linear model. Possibilities include perceptron training (Collins, 2002), log-linear optimisation of the conditional log-likelihood (Berger, Pietra, & Pietra, 1996) and large margin methods. We base our training on Tsochantaridis et al.'s (2005) framework for learning Support Vector Machines (SVMs) over structured output spaces, using the SVM$^{struct}$ implementation.[9] The framework supports a configurable loss function which is particularly appealing in the context of sentence compression and more generally text-to-text generation. It also has an efficient training algorithm and powerful regularization. The latter is is critical for discriminative models with large numbers of features, which would otherwise over-fit the training sample at the expense of generalization accuracy. We briefly summarize the approach below; for a more detailed description we refer the interested reader to the work of Tsochantaridis et al. (2005).

Traditionally SVMs learn a linear classifier that separates two or more classes with the largest possible margin. Analogously, structured SVMs attempt to separate the correct structure from all other structures with a large margin. The learning objective for the structured SVM uses the soft-margin formulation which allows for errors in the training set via the slack variables, $\xi_i$:

$$\min_{\mathbf{w}, \xi} \frac{1}{2} ||\mathbf{w}||^2 + \frac{\mathcal{C}}{n} \sum_{i=1}^{n} \xi_i, \; \xi_i \geq 0 \tag{10}$$

$$\forall i, \forall \mathbf{d} : source(\mathbf{d}) = \mathbf{x}_i \land \mathbf{y} \neq \mathbf{d}_i : \langle \mathbf{w}, \Psi(\mathbf{d}_i) - \Psi(\mathbf{d}) \rangle \geq \Delta(\mathbf{d}_i, \mathbf{d}) - \xi_i$$

The slack variables, $\xi_i$, are introduced here for each training example, $\mathbf{x}_i$ and $\mathcal{C}$ is a constant that controls the trade-off between training error minimization and margin maximization. Note that slack variables are combined with the loss incurred in each of the linear constraints. This means that a high loss output must be separated by a larger margin than a low loss output, or have a much larger slack variable to satisfy the constraint. Alternatively, the loss function can be used to rescale the slack parameters, in which case the constraints in (10) are replaced with $\langle \mathbf{w}, \Psi(\mathbf{d}_i) - \Psi(\mathbf{d}) \rangle \geq 1 - \frac{\xi_i}{\Delta(\mathbf{d}_i, \mathbf{d})}$. Margin rescaling is theoretically less desirable as it is not scale invariant, and therefore requires the tuning of an additional hyperparameter compared to slack rescaling. However, empirical results show little difference between the two rescaling methods (Tsochantaridis et al., 2005). We use margin rescaling for the practical reason that it can be approximated more accurately than can slack rescaling by our chart based inference method.

---

9. `http://svmlight.joachims.org/svm_struct.html`

The optimization problem in (10) is approximated using an algorithm proposed by Tsochantaridis et al. (2005). The algorithm finds a small set of constraints from the full-sized optimization problem that ensures a sufficiently accurate solution. Specifically, it constructs a nested sequence of successively tighter relaxation of the original problem using a (polynomial time) cutting plane algorithm. For each training instance, the algorithm keeps track of the selected constraints defining the current relaxation. Iterating through the training examples, it proceeds by finding the output that most radically violates a constraint. In our case, the optimization crucially relies on finding the derivation which is both high scoring and has high loss compared to the gold standard. This requires finding the maximizer of:

$$H(\mathbf{d}) = \Delta(\mathbf{d}^*, \mathbf{d}) - \langle \mathbf{w}, \Psi(\mathbf{d}_i) - \Psi(\mathbf{d}) \rangle \tag{11}$$

The search for the maximizer of $H(\mathbf{d})$ in (11) can be performed by the decoding algorithm presented in Section 3.4 with some extensions. Firstly, by expanding (11) to $H(\mathbf{d}) = \Delta(\mathbf{d}^*, \mathbf{d}) - \langle \Psi(\mathbf{d}_i), \mathbf{w} \rangle + \langle \Psi(\mathbf{d}), \mathbf{w} \rangle$ we can see that the second term is constant with respect to $\mathbf{d}$, and thus does not influence the search. The decoding algorithm maximizes the last term, so all that remains is to include the loss function into the search process.

Loss functions which decompose with the rules or target ngrams in the derivation, $\Delta(\mathbf{d}^*, \mathbf{d}) = \sum_{r \in \mathbf{d}} \Delta_R(\mathbf{d}^*, r) + \sum_{n \in ngrams(\mathbf{d})} \Delta_N(\mathbf{d}^*, n)$, can be easily integrated into the decoding algorithm. This is done by adding the partial loss, $\Delta_R(\mathbf{d}^*, r) + \Delta_N(\mathbf{d}^*, n)$ to each rule's score in line 20 of Algorithm 4 (the ngrams are recovered from the ngram contexts in the same manner used to evaluate the ngram features).

However, many of our loss functions do not decompose with the rules or the ngrams. In order to calculate these losses the chart must be *stratified* by the loss function's *arguments* (Joachims, 2005). For example, unigram precision measures the ratio of correctly predicted tokens to total predicted tokens and therefore its loss arguments are the pair of counts, $(TP, FP)$, for true and false positives. They are initialized to $(0, 0)$ and are then updated for each rule used in a derivation. This equates to checking whether each target terminal is in the reference string and incrementing the relevant value. The chart is extended (stratified) to store the loss arguments in the same way that ngram contexts are stored for decoding. This means that a rule accessing a child chart cell can get multiple entries, each with different loss argument values as well as multiple ngram contexts (line 14 in Algorithm 4). The loss argument for a rule application is calculated from the rule itself and the loss arguments of its children. This is then stored in the chart and the back-pointer list (lines 22–23 in Algorithm 4). Although this loss can only be evaluated correctly for complete derivations, we also evaluate the loss on partial derivations as part of the cube-pruning heuristic. Losses with a large space of argument values will be more coarsely approximated by the beam search, which prunes the number of chart entries to a constant size. For this reason, we have focused mainly on simple loss functions which have a relatively small space of argument values, and also use a wide beam during the search (200 unique items or 500 items, whichever comes first).

---

**Algorithm 5** Find the gold standard derivation for a pair of trees (i.e., alignment).

---

**Require:** source tree, $\mathbf{x}$, and target tree, $\mathbf{y}$

1: let $C[v_S, v_T] \in \mathcal{R}$ be a chart representing the maximum number of rules used to align nodes $v_S \in \mathbf{x}$ and $v_T \in \mathbf{y}$

2: let $B[v_S, v_T] \in (P, \mathbf{x} \times \mathbf{y})$ be the corresponding back-pointers, consisting of a production and a pair aligned nodes for each of the production's variables

3: initialize chart, $C[*, *] = -\infty$

4: initialize back-pointers, $B[*, *] = $ none

5: **for all** source nodes, $v_S \in \mathbf{x}$, bottom-up **do**

6:   **for all** rules, $r = \langle v_S, Y \rangle \rightarrow \langle \alpha, \gamma, \sim \rangle$ where $\alpha$ matches the sub-tree rooted at $v_S$ **do**

7:     **for all** target nodes, $v_T \in \mathbf{y}$, matching $\gamma$ **do**

8:       let rule count, $j \leftarrow 1$

9:       **for all** variables, $u \in \sim$ **do**

10:         find aligned child nodes, $(c_S, c_T)$, under $v_S$ and $v_T$ corresponding to $u$

11:         update rule count, $j \leftarrow j + C[c_S, c_T]$

12:       **end for**

13:       **if** $n$ greater than previous value in chart **then**

14:         update chart, $C[v_S, v_T] \leftarrow j$

15:         update back-pointers, $B[v_S, v_T] \leftarrow (r, \{(c_S, c_T) \forall u\})$

16:       **end if**

17:     **end for**

18:   **end for**

19: **end for**

20: **if** $C[root(\mathbf{x}), root(\mathbf{y})] \neq -\infty$ **then**

21:   success; create derivation by traversing back-pointers from $B[root(\mathbf{x}), root(\mathbf{y})]$

22: **end if**

---

In our discussion so far we have assumed that we are given a gold standard derivation, $\mathbf{y}_i$ glossing over the issue of how to find it. Spurious ambiguity in the grammar means that there are often many derivations linking the source and target, none of which are clearly 'correct'. We select the derivation using the maximum number of rules, each of which will be small, and therefore should provide maximum generality.[10] This is found using Algorithm 5, a chart-based dynamic program similar to the alignment algorithm for inverse transduction grammars (Wu, 1997). The algorithm has time complexity $O(S^2 R)$ where $S$ is the size of the larger of the two trees and $R$ is the number of rules which can match a node.

### 3.6 Loss Functions

The training algorithm described above is highly modular and in theory can support a wide range of loss functions. There is no widely accepted evaluation metric for text compression. A zero-one loss would be straightforward to define but inappropriate for our problem, as it would always penalize target derivations that differ even slightly from the reference derivation. Ideally, we would like a loss with a wider scoring range that can discriminate between derivations that differ from the reference. Some of these may be good compressions whereas others may be entirely ungrammatical. For this reason we have developed a range of loss functions which draw inspiration from various metrics used for evaluating text-to-text rewriting tasks such as summarization and machine translation.

---

10. We also experimented with other heuristics, including choosing the derivation at random and selecting the derivation with the maximum or minimum score under the model (all using the same search algorithm but with a different objective). Of these, only the maximum scoring derivation was competitive with the maximum rules heuristic.

Loss functions are defined over derivations and can look at any *item* accessible including tokens, ngrams and CFG rules. Our first class of loss functions calculates the Hamming distance between unordered bags of items. It measures the number of predicted items that did not appear in the reference, along with a penalty for short output:

$$\Delta_{hamming}(\mathbf{d}^*, \mathbf{d}) = FP + \max\left(l - (TP + FP), 0\right) \tag{12}$$

where $TP$ and $FP$ are the number of true and false positives, respectively, when comparing the predicted target, $\mathbf{d}_T$, with the reference, $\mathbf{d}_T^*$, and $l$ is the length of the reference. We include the second term to penalize overly short output as otherwise predicting very little or nothing would incur no penalty.

We have created three instantiations of the loss function in (12) over: 1) tokens, 2) ngrams ($n \leq 3$), and 3) CFG productions. In each case, the loss argument space is quadratic in the size of the source tree. Our Hamming ngram loss is an attempt at defining a loss function similar to BLEU (Papineni, Roukos, Ward, & Zhu, 2002). The latter is defined over documents rather than individual sentences, and is thus not directly applicable to our problem. Now, since these losses all operate on unordered bags they may reward erroneous predictions, for example, a permutation of the reference tokens will have zero token-loss. This is less of a problem for the CFG and ngram losses whose items overlap, thereby encoding a partial order. Another problem with the loss functions just described is that they do not penalize multiply predicting an item that occurred only once in the reference. This could be a problem for function words which are common in most sentences.

Therefore we developed two additional loss functions which take multiple predictions into account. The first measures the edit distance — the number of insertions and deletions — between the predicted and the reference compressions, both as bags-of-tokens. In contrast to the previous loss functions, it requires the true positive counts to be clipped to the number of occurrences of each type in the reference. The edit distance is given by:

$$\Delta_{edit}(\mathbf{d}^*, \mathbf{d}) = p + r - 2\sum_i \min(p_i, q_i) \tag{13}$$

where $p$ and $q$ denote the number of target tokens in the predicted tree, $target(\mathbf{d})$, and reference, $\mathbf{y}^* = target(\mathbf{d}^*)$, respectively, and $p_i$ and $q_i$ are the counts for type $i$. The loss arguments for the edit distance consist of a vector of counts for each item type in the reference, $\{p_i, \forall i\}$. The space of possible values is exponential in the size of the source tree, compared to quadratic for the Hamming losses. Consequently, we expect beam search to result in many more search errors when using the edit distance loss.

Our last loss function is the F1 measure, a harmonic mean between precision and recall, measured over bags-of-tokens. As with the edit distance, its calculation requires the counts to be clipped to the number of occurrences of each terminal type in the reference. We

**Ref:**  [S [WHNP [WP what]] [S [NP [NNS records]] [VP [VBP are] [VP [VBN involved]]]]]]
**Pred:**  [S [WHNP [WP what]] [S [NP [NNS ones]] [VP [VBP are] [VBN involved]]]]]

| Loss | Arguments | Value |
|---|---|---|
| Token Hamming | $TP = 3, FP = 1$ | 1/4 |
| 3-gram Hamming | $TP = 8, FP = 5$ | 5/14 |
| CFG Hamming | $TP = 8, FP = 1$ | 1/9 |
| Edit distance | $p = (1, 0, 1, 1, 1)$ | 2 |
| F1 | $p = (1, 0, 1, 1, 1)$ | 1/4 |

Table 1: Loss arguments and values for the example predicted and reference compressions. Note that loss values should not be compared between different loss functions; these values are purely illustrative.

therefore use the same loss arguments for its calculation. The $F_1$ loss is given by:

$$\Delta_{F_1}(\mathbf{d}^*, \mathbf{d}) = 1 - \frac{2 \times precision \times recall}{precision + recall} \tag{14}$$

where $precision = \frac{\sum_i min(p_i, q_i)}{p}$ and $recall = \frac{\sum_i min(p_i, q_i)}{q}$. As F1 shares the same arguments with the edit distance loss, it also has the same exponential space of loss argument values and will consequently be subject to severe pruning during the beam search used in training.

To illustrate the above loss functions, we present an example in Table 1. Here, the prediction (Pred) and reference (Ref) have the same length (4 tokens), identical syntactic structure, but differ by one word (*ones* versus *records*). Correspondingly, there are three correct tokens and one incorrect, which forms the arguments for the token Hamming loss, resulting in a loss of 1/4. The ngram loss is measured for $n \leq 3$ and the start and end of the string are padded with special symbols to allow evaluation of the boundary ngrams. The CFG loss records only one incorrect CFG production (the preterminal [NNS ones]) from the total of nine productions. The last two losses use the same arguments: a vector with values for the counts of each reference type. The first four cells correspond to *what*, *records*, *are* and *involved*, the last cell records all other types. For the example, the edit distance is two (one deletion and one insertion) while the F1 loss is 1/4 (precision and recall are both 3/4).

## 4. Features

Our feature space is defined over source trees, $\mathbf{x}$, and target derivations, $\mathbf{d}$. We devised two broad classes of features, applying to grammar rules and to ngrams of target terminals. We defined only a single ngram feature, the conditional log-probability of a trigram language model. This was trained on the BNC (100 million words) using the SRI Language Modeling toolkit (Stolcke, 2002), with modified Kneser-Ney smoothing.

For each rule $\langle X, Y \rangle \to \langle \alpha, \gamma, \sim \rangle$, we extract features according to the templates detailed below. Our templates give rise to binary indicator features, except where explicitly stated. These features perform a boolean test, returning value 1 when the test succeeds and 0 otherwise. An example rule and its corresponding features are shown in Table 2.

**Type**: Whether the rule was extracted from the training set, created as a copy rule and/or created as a delete rule. This allows the model to learn a preference for each of the three sources of grammar rules (see row Type in Table 2)

**Root**: The root categories of the source, $X$, and target, $Y$, and their conjunction, $X \wedge Y$ (see rows Root in Table 2).

**Identity**: The source side, $\alpha$, target side, $\gamma$, and the full rule, $(\alpha, \gamma, \sim)$. This allows the model to learn weights on individual rules or those sharing an elementary tree. Another feature checks if the rule's source and target elementary trees are identical, $\alpha = \gamma$ (see rows Identity in Table 2).

**Unlexicalised Identity**: The identity feature templates above are replicated for unlexicalised elementary trees, i.e., with the terminals removed from their frontiers (see rows UnlexId in Table 2).

**Rule count**: This feature is always 1, allowing the model to count the number of rules used in a derivation (see row Rule count in Table 2).

**Word count**: Counts the number of terminals in $\gamma$, allowing a global preference for shorter or longer output. Additionally, we record the number of terminals in the source tree, which can be used with the target terminal count to find the number of deleted terminals (see rows Word count in Table 2).

**Yield**: These features compare the terminal yield of the source, $Y(\alpha)$, and target, $Y(\gamma)$. The first feature checks the identity of two sequences, $Y(\alpha) \wedge Y(\gamma)$. We use identity features for each terminal in both yields, and for each terminal only in the source (see rows Yield in Table 2). We also replicate these feature templates for the sequence of non-terminals on the frontier (pre-terminals or variable non-terminals).

**Length**: Records the difference in the lengths of the frontiers of $\alpha$ and $\gamma$, and whether the target's frontier is shorter than that of the source (see rows Length in Table 2).

The features listed above are defined for all the rules in the grammar. This includes the copy and delete rules, as described in Section 3.2, which were added to address the problem of unseen words or productions in the source trees at test time. Many of these rules can not be applied to the training set, but will receive some weight because they share features with rules that can be used in training. However, in training the model learns to disprefer these coverage rules as they are unnecessary to model the training set, which can be described perfectly using the extracted transduction rules. Our dual use of the training set for grammar extraction and parameter estimation results in a bias against the coverage rules. The bias could be addressed by extracting the grammar from a separate corpus, in which case the coverage rules would then be useful in modeling both the training set and the testing sets. However, this solution has its own problems, namely that many of the target trees in the training may not longer be reachable. This bias and its possible solutions is an interesting research problem and deserves further work.

| Rule: ⟨NP,NNS⟩ → ⟨[NP CD☐ ADJP☐ [NNS activists]], [NNS activists]⟩ | | |
|---|---|---|
| Type | type = training set | 1 |
| Root | $X$ = NP | 1 |
| Root | $Y$ = NNS | 1 |
| Root | $X$ = NP ∧ $Y$ = NNS | 1 |
| Identity | $\alpha$ = [NP CD ADJP [NNS activists]] | 1 |
| Identity | $\gamma$ = [NNS activists] | 1 |
| Identity | $\alpha$ = [NP CD☐ ADJP☐ [NNS activists]] ∧ $\gamma$ = [NNS activists] | 1 |
| UnlexId. | unlex. $\alpha$ = [NP CD ADJP NNS] | 1 |
| UnlexId. | unlex. $\gamma$ = NNS | 1 |
| UnlexId. | unlex. $\alpha$ = [NP CD☐ ADJP☐ NNS] ∧ $\gamma$ = NNS | 1 |
| Rule count | – | 1 |
| Word count | target terminals | 1 |
| Word count | source terminals | ≥ 1* |
| Yield | source = ['activists'] ∧ target = ['activists'] | 1 |
| Yield | terminal 'activists' in both source and target | 1 |
| Yield | non-terms. source = [CD, ADJP, NNS] ∧ target = [NNS] | 1 |
| Yield | non-terminal CD in source and not target | 1 |
| Yield | non-terminal ADJP in source and not target | 1 |
| Yield | non-terminal NNS in both source and target | 1 |
| Length | difference in length | 2 |
| Length | target shorter | 1 |

Table 2: Features instantiated for the synchronous rule shown above. Only features with non-zero values are displayed. *The number of source terminals is calculated using the source tree at the time the rule is applied.

## 5. Experimental Set-up

In this section we present our experimental set-up for assessing the performance of the sentence compression model described above. We give details of the corpora used, briefly introduce McDonald's (2006) model used for comparison with our approach, and explain how system output was evaluated.

### 5.1 Corpora

We evaluated our system on three publicly available corpora. The first is the Ziff-Davis corpus, a popular choice in the sentence compression literature. The corpus originates from a collection of news articles on computer products. It was created automatically by matching sentences that occur in an article with sentences that occur in an abstract (Knight & Marcu, 2002). The other two corpora[11] were created manually; annotators were asked to produce target compressions by deleting extraneous words from the source without changing the word order (Clarke & Lapata, 2008). One corpus was sampled from written sources, the British National Corpus (BNC) and the American News Text corpus, whereas the other was created from manually transcribed broadcast news stories. We will henceforth refer to these two corpora as CLwritten and CLspoken, respectively. The sizes of these three corpora are shown in Table 3.

---

11. Available from `http://homepages.inf.ed.ac.uk/s0460084/data/`.

| Corpus | Articles | Sentences | Training | Development | Testing |
|--------|----------|-----------|----------|-------------|---------|
| CLspoken | 50 | 1370 | 882 | 78 | 410 |
| CLwritten | 82 | 1433 | 908 | 63 | 462 |
| Ziff-Davis | – | 1084 | 1020 | 32 | 32 |

Table 3: Sizes of the various corpora, measured in articles or sentence pairs. The data split into training, development and testing sets is measured in sentence pairs.

These three corpora pose different challenges to a hypothetical sentence compression system. Firstly, they are representative of different domains and text genres. Secondly, they have different compression requirements. The Ziff-Davis corpus is more aggressively compressed in comparison to CLspoken and CLwritten (Clarke & Lapata, 2008). As CLspoken is a speech corpus, it often contains incomplete and ungrammatical utterances and speech artefacts such as disfluencies, false starts and hesitations. Its utterances have varying lengths, some are very wordy whereas others cannot be reduced any further. This means that a compression system should leave some sentences uncompressed. Finally, we should note the CLwritten has on average longer sentences than Ziff-Davis or CLspoken. Parsers are more likely to make mistakes on long sentences which could potentially be problematic for syntax-based systems like the one presented here.

Although our model is capable of performing any editing operation, such as reordering or substitution, it will not learn to do so from the training corpora. These corpora contain only deletions, and therefore the model will not learn transduction rules encoding, e.g., reordering. Instead the rules encode only the deleting and inserting terminals and restructuring internal nodes of the syntax tree. However, the model is capable general text rewriting, and given the appropriate training set will learn to perform these additional edits. This is demonstrated by our recent results from adapting the model to *abstractive compression* (Cohn & Lapata, 2008), where any edit is permitted, not just deletion.

Our experiments on CLspoken and CLwritten followed Clarke and Lapata's (2008) partition of training, test, and development sets. The partition sizes are shown in Table 3. In the case of the Ziff-Davis corpus, Knight and Marcu (2002) had not defined a development set. Therefore we randomly selected (and held-out) 32 sentence pairs from their training set to form our development set.

## 5.2 Comparison with State-of-the-Art

We evaluated our results against McDonald's (2006) discriminative model. In this approach, sentence compression is formalized as a classification task: pairs of words from the source sentence are classified as being adjacent or not in the target compression. Let $\mathbf{x} = x_1, \ldots, x_N$ denote a source sentence with a target compression $\mathbf{y} = y_1, \ldots, y_M$ where each $y_i$ occurs in $\mathbf{x}$. The function $L(y_i) \in \{1 \ldots N\}$ maps word $y_i$ the target to the index of the word in the source (subject to the constraint that $L(y_i) < L(y_{i+1})$). McDonald defines the score of a compression $\mathbf{y}$ for a sentence $\mathbf{x}$ as the dot product between a high dimensional feature representation, $\mathbf{f}$, over bigrams and a corresponding weight vector, $\mathbf{w}$,

$$score(\mathbf{x}, \mathbf{y}; \mathbf{w}) = \sum_{i=2}^{M} \langle \mathbf{w}, \mathbf{f}(\mathbf{x}, L(y_{j-1}), L(y_j)) \rangle \qquad (15)$$

Decoding in this framework amounts to finding the combination of bigrams that maximize the scoring function in (15). The maximization is solved using a semi-Markov Viterbi algorithm (McDonald, 2006).

The model parameters are estimated using the Margin Infused Relaxed Algorithm (MIRA Crammer & Singer, 2003), a discriminative large-margin online learning technique. McDonald (2006) uses a similar loss function to our Hamming loss (see (12)) but without an explicit length penalty. This loss function counts the number of words falsely retained or dropped in the predicted target relative to the reference. McDonald employs a rich feature set defined over words, parts of speech, phrase structure trees, and dependencies. These are gathered over adjacent words in the compression and the words which were dropped.

Clarke and Lapata (2008) reformulate McDonald's (2006) model in the context of integer linear programming (ILP) and augment it with constraints ensuring that the compressed output is grammatically and semantically well formed. For example, if the target sentence has negation, this must be included in the compression; If the source verb has a subject, this must also be retained in the compression. They generate and solve an ILP for every source sentence using the branch-and-bound algorithm. Since they obtain performance improvements over McDonald's model on several corpora, we also use it for comparison against our model.

To summarize, we believe that McDonald's (2006) model is a good basis for comparison for several reasons. First, it is has good performance, and can be treated as a state-of-the-art model. Secondly, it is similar to our model in many respects – its training algorithm and feature space – but differs in one very important respect: compression is performed on strings and not trees. McDonald's system does make use of syntax trees, but only peripherally via the feature set. In contrast, the syntax tree is an integral part of our model.

## 5.3 Evaluation

In line with previous work we assessed our model's output by eliciting human judgments. Following Knight and Marcu (2002), we conducted two separate experiments. In the first experiment participants were presented with a source sentence and its target compression and asked to rate how well the compression preserved the most important information from the source sentence. In the second experiment, they were asked to rate the grammaticality of the compressed outputs. In both cases they used a five point rating scale where a high number indicates better performance. We randomly selected 20 sentences from the test portion of each corpus. These sentences were compressed automatically by our system and McDonald's (2006) system. We also included gold standard compressions. Our materials thus consisted of 180 ($20 \times 3 \times 3$) source-target sentences. A Latin square design ensured that subjects did not see two different compressions of the same sentence. We collected ratings from 30 unpaid volunteers, all self reported native English speakers. Both studies were conducted over the Internet using WebExp,[12] a software package for running Internet-based experiments.

We also report results using F1 computed over grammatical relations (Riezler et al., 2003). Although F1 conflates grammaticality and importance into a single score, it nevertheless has been shown to correlate reliably with human judgments (Clarke & Lapata, 2006). Furthermore, it can be usefully employed during development for feature engineering and parameter optimization experiments. We measured F1 over directed and labeled dependency relations. For all models the compressed output was parsed using the RASP dependency parser (Briscoe & Carroll, 2002). Note that we could extract dependencies directly from the output of our model since it generates trees in addition to strings. However, we refrained from doing this in order to compare all models on an equal footing.

# 6. Results

The framework presented in Section 3 is quite flexible. Depending on the grammar extraction strategy, choice of features, and loss function, different classes of models can be derived. Before presenting our results on the test set we discuss the specific model employed in our experiments and explain how its parameters were instantiated.

## 6.1 Model Selection

All our parameter tuning and model selection experiments were conducted on the development set of the CLspoken corpus. We obtained syntactic analyses for source and target sentences with Bikel's (2002) parser. The corpus was automatically aligned using an algorithm which finds the set of deletions which transform the source into the target. This is equivalent to the minimum edit distance script when only deletion operations are permitted.

As expected, the predicted parse trees contained a number of errors, although we did not have gold standard trees with which to quantify this error or its effect on prediction output. We did notice, however, that errors in the source trees in the test set did not always negatively affect the performance of the model. In many instances the model was able to recover from these errors and still produce good output compressions. Of these recoveries, most cases involved either deleting the erroneous structure or entirely preserving it. While this often resulted in a poor output tree, the string yield was acceptable in most cases. Less commonly, the model corrected the errors in the source using tree transformation rules. These rules were acquired from the training set where there were errors in the source tree but not in the test tree. For example, one transformation allows a prepositional phrase to be moved from a high VP attachment to an object NP attachment.

We obtained a synchronous tree substitution grammar from the CLspoken corpus using the method described in Section 3.2. We extracted all maximally general synchronous rules. These were complemented with specified rules allowing recursion up to one ancestor for any given node.[13] Grammar rules were represented by the features described in Section 4. An important parameter in our modeling framework is the choice of loss function. We

---

| Losses | Rating | Std. dev |
|---|---|---|
| Hamming (tokens) | 3.38 | 1.05 |
| Hamming (ngram) | 3.28 | 1.13 |
| Hamming (CFG) | 3.22 | 0.91 |
| Edit Distance | 3.30 | 1.20 |
| F1 | 3.15 | 1.13 |
| Reference | 4.28 | 0.70 |

Table 4: Mean ratings on system output (CLspoken development set) while using different loss functions.

evaluated the loss functions presented in Section 3.6 as follows. We performed a grid search for the hyper-parameters (a regularization parameter and a feature scaling parameter, which balances the magnitude of the feature vectors with the scale of the loss function)[14] which minimized the relevant loss on the development set, and used the corresponding system output. The gold standard derivation was selected using the maximum number of rules heuristic, as described in Section 3.5. The beam was limited to 100 unique items or 200 items in total. The grammar was filtered to allow no more than 50 target elementary trees for every source elementary tree.

We next asked two human judges to rate on a scale of 1 to 5 the system's compressions when optimized for the different loss functions. To get an idea of the quality of the output we also included human-authored reference compressions. Sentences given high numbers were both grammatical and preserved the most important information. The mean ratings are shown in Table 4. As can be seen the differences among the losses are not very large, and the standard deviation is high. The Hamming loss over tokens performed best with a mean rating of 3.38, closely followed by the edit distance (3.30). We chose the former over the latter as it is less coarsely approximated during search. All subsequent experiments report results using the token-based Hamming loss.

We also wanted to investigate how the synchronous grammar influences performance. The default system described above used general rules together with specialized rules where the recursion depth was limited to one. We also experimented with a grammar that uses specialised rules with a maximum recursion depth of two and a grammar that uses solely the maximally general rules. In Table 5 we report the average compression rate, relations-based F1 and the Hamming loss over tokens for these different grammars. We see that adding the specified rules allows for better F1 (and loss) despite the fact that the search space remains the same. We observe a slight degradation in performance moving to depth $\leq 2$ rules. This is probably due to the increase in spurious ambiguity affecting search quality, and also allowing greater overfitting of the training data. The number of transduction rules in the grammar also grows substantially with the increased depth – from 20,764 for the maximally general extraction technique to 33,430 and 62,116 for specified rules with depth $\leq 1$ and $\leq 2$, respectively. The growth in grammar size is exponential in the specification depth and therefore only small values should be used.

---

14. We found that setting the regularization parameter $\mathcal{C} = 0.01$ and the scaling parameter to 1 generally yields good performance across loss functions.

| Model | Compression rate | Relations F1 | Loss |
|---|---|---|---|
| max general rules | 80.79 | 65.04 | 341 |
| depth $\leq$ 1-specified rules* | 79.72 | 68.56 | 315 |
| depth $\leq$ 2-specified rules | 79.71 | 66.44 | 328 |
| max rules* | 79.72 | 68.56 | 315 |
| max scoring | 81.03 | 65.54 | 344 |
| unigram LM | 76.83 | 59.05 | 336 |
| bigram LM | 83.12 | 67.71 | 317 |
| trigram LM* | 79.72 | 68.56 | 315 |
| all features* | 79.72 | 68.56 | 315 |
| only rule features | 83.06 | 67.51 | 346 |
| only token features | 85.10 | 68.31 | 341 |

Table 5: Parameter exploration and feature ablation studies (CLspoken development set). The default system is shown with an asterisk.

We also inspected the rules obtained with the maximally general extraction technique to better assess how our rules differ from those obtained from a vanilla SCFG (see Knight & Marcu, 2002). Many of these rules (12%) have deeper structure and therefore would not be licensed by an SCFG. This is due to structural divergences between the source and target syntax trees in the training set. A further 13% of the rules describe a change of syntactic category ($X \neq Y$), and therefore only the remaining 76% of the rules would be allowable in Knight and Marcu's transducer. The proportion of SCFG rules decreases substantially as the rule specification depth is increased.

Recall from Section 3.3 that our scoring function is defined over derivations rather than target trees or strings, and that we treat the derivation using the maximum number of rules as the gold standard derivation. As a sanity check, we also experimented with selecting the derivation with the maximum score under the model. The results in Table 5 indicate that the latter strategy is not as effective as selecting the derivation with the maximum number of rules. Again we conjecture this is due to overfitting. As the training data is used to extract the grammar, the derivations with the maximum score may consist of rules with rare features which model the data well but do not generalize to unseen instances.

Finally, we conducted a feature ablation study to assess which features are more useful to our task. We were particularly interested to see if the ngram features would bring any benefit, especially since they increase computational complexity during decoding and training. We experimented with a unigram, bigram, and trigram language model. Note that the unigram language model is not as computationally expensive as the other two models because there is no need to record ngram contexts in the chart. As shown in Table 5, the unigram language model is substantially worse than the bigram and trigram which deliver similar performances. We also examined the impact of the other features by grouping them into two broad classes, those defined over rules and those defined over tokens. Our aim was to see whether the underlying grammar (represented by rule-based features) contributes to better compression output. The results in Table 5 reveal that the two feature groups perform comparably. However, the model using only token-based features tends to compress less. These features are highly lexicalized, and the model is not able to generalize well on unseen data. In conclusion, the full feature set does better on all counts than the two ablation sets, with a better compression rate.

The results reported have all been measured over string output. This was done by first stripping the tree structure from the compression output, reparsing, extracting dependency relations and finally comparing to the dependency relations in the reference. However, we may wish to measure the quality of the trees themselves, not just their string yield. A simple way to measure this[15] would be to extract dependency relations directly from the phrase-structure tree output.[16] Compared to dependencies extracted from the predicted parses using Bikel's (2002) parser on the output string, we observe that the relation F1 score increases uniformly for all tasks, by between 2.50% and 4.15% absolute. Therefore the system's tree output better encodes the syntactic dependencies than the tree resulting from re-parsing the string output. If the system is part of a NLP pipeline, and its output is destined for down-stream processing, then having an accurate syntax tree is extremely important. This is also true for related tasks where the desired output is a tree, e.g., semantic parsing.

## 7. Model Comparison

In this section we present our results on the test set using the best performing model from the previous section. This model uses a grammar with unlexicalized and lexicalized rules (recursion depth 1), a Hamming loss based on tokens, and all the features from Section 4. The model was trained separately on each corpus (training portion). We first discuss our results using relations F1 and then move on to the human study.

Table 6 illustrates the performance of our model (Transducer1) on CLspoken, CLwritten, and Ziff Davis. We also report results on the same corpora using McDonald's (2006) model (McDonald) and the improved version (Clarke ILP) put forward by Clarke and Lapata (2008). We also present the compression rate for each system and the reference gold standard. In all cases our tree transducer model outperforms McDonald's original model and the improved ILP-based version.

Nevertheless, it may be argued that our model has an unfair advantage here since it tends to compress less than the other models, and is therefore less likely to make many mistakes. To ensure that this is not the case, we created a version of our model with a compression rate similar to McDonald. This can be done relatively straightforwardly by manipulating the length penalty of the Hamming loss. The smaller the penalty the more words the model will tend to drop. Therefore, we varied the length penalty (and hyper-parameters) on the development set in order to obtain a compression rate similar to McDonald.[17] This model was then applied to the test set and its performance is shown in Table 6 as Transducer2. We refrained from doing this on Ziff-Davis, since our original transducer obtained a compression rate comparable to McDonald (67.45 vs. 66.26). As can be seen, Transducer2 yields a better F1 on CLspoken and CLwritten. The differences in F1 are statistically significant using the the Wilcoxon test ($p < 0.01$). Transducer1 numerically outperforms McDonald on Ziff-Davis, however the difference is not significant (the Ziff-Davis test set consists solely of 32 sentences).

---

15. We could alternatively measure other tree metrics, such as tree edit distance. However, the standard measures used in parser evaluation (e.g., EVALB) would not be suitable, as they assume that the parse yield is fixed. In our case the reference target string is often different to the system's output.

16. We extract dependency relations with the conversion tool from the CoNLL 2007 shared task, available at `http://nlp.cs.lth.se/pennconverter/`.

| CLspoken | | |
| --- | --- | --- |
| Model | Compression rate | Relations F1 |
| Transducer1 | 82.30 | 66.63 |
| Transducer2 | 69.89 | 59.58 |
| McDonald | 68.56 | 47.48 |
| Clarke ILP | 77.70 | 54.12 |
| Reference | 76.11 | – |

| CLwritten | | |
| --- | --- | --- |
| Model | Compression rate | Relations F1 |
| Transducer1 | 76.52 | 58.02 |
| Transducer2 | 61.09 | 49.48 |
| McDonald | 60.12 | 48.39 |
| Clarke ILP | 71.99 | 54.84 |
| Reference | 70.24 | – |

| Ziff Davis | | |
| --- | --- | --- |
| Model | Compression rate | Relations F1 |
| Transducer1 | 67.45 | 56.55 |
| McDonald | 66.26 | 54.12 |
| Clarke ILP | 48.67 | 46.77 |
| Reference | 56.61 | – |

Table 6: Results on CLspoken, CLwritten, and Ziff Davis corpus (testing set); compression rate and relations-based F1.

We next consider the results of our judgment elicitation study which assesses in more detail the quality of the generated compressions. Recall that our participants judge compressed output on two dimensions, grammaticality and importance. We compared the output of our system (Transducer2 on CLspoken and CLwritten and Transducer1 on Ziff-Davis) against the output of McDonald (2006) and the reference gold standard. Table 7 illustrates examples of the compressions our participants saw.

---

17. We matched the compression rate of McDonald by scaling the length penalty by 0.50 and 0.25 for the CLwritten and CLspoken corpora, respectively. Another way to control the compression rate would be to modify our chart-based decoder in a fashion similar to McDonald (2006). However, we leave this to future work.

S: I just wish my parents and my other teachers could be like this teacher, so we could communicate.
M: I wish my teachers could be like this teacher.
T: I wish my teachers could be like this, so we could communicate.
R: I wish my parents and other teachers could be like this, so we could communicate.

S: The Treasury is refusing to fund a further phase of the city technology colleges.
M: The Treasury is refusing to fund a further colleges.
T: The Treasury is refusing to fund the city technology colleges.
R: The Treasury is refusing to fund further the city technology colleges.

S: Apparel makers use them to design clothes and to quickly produce and deliver the best-selling garments.
M: Apparel makers use them to design clothes and to produce and deliver the best-selling garments.
T: Apparel makers use them to design clothes.
R: Apparel makers use them to design clothes.

S: Earlier this week, in a conference call with analysts, the bank said it boosted credit card reserves by $350 million.
M: Earlier said credit card reserves by $350 million.
T: In a conference call with analysts, the bank boosted card reserves by $350 million.
R: In a conference call with analysts the bank said it boosted credit card reserves by $350 million.

Table 7: Compression examples from CLspoken, CLwritten, and Ziff-Davis (S: source sentence, M: McDonald, 2006, T: transducer, R: reference gold standard)

Table 8 shows the mean ratings[18] for each system (and the reference) on CLspoken, CLwritten, and Ziff-Davis. We carried out an Analysis of Variance (ANOVA) to examine the effect of system type (MCDONALD, TRANSDUCER, Reference) on the compression ratings. The ANOVA revealed a reliable effect on all three corpora. We used post-hoc Tukey tests to examine whether the mean ratings for each system differed significantly ($p < 0.01$). On the CLspoken corpus the TRANSDUCER is perceived as significantly better than MCDONALD, both in terms of grammaticality and importance. We obtain the same result for the CLwritten corpus. The two systems achieve similar performances on Ziff-Davis (the grammaticality and importance score do not differ significantly). Ziff-Davis seems to be a less challenging corpus than CLspoken or CLwritten and less likely to highlight differences among systems. For example, Turner and Charniak (2005) present several variants of the noisy-channel model, all of which achieve compressions of similar quality on Ziff-Davis (grammaticality ratings varied by only $\pm0.13$ and informativeness ratings $\pm0.31$ in their human evaluation). In most cases the TRANSDUCER and MCDONALD yield significantly

---

18. All statistical tests reported subsequently were done using the mean ratings.

| CLspoken | | |
|---|---|---|
| Model | Grammaticality | Importance |
| TRANSDUCER | 4.18* | 3.98* |
| McDONALD | 2.74$^\dagger$ | 2.51$^\dagger$ |
| Reference | 4.58 | 4.22 |

| CLwritten | | |
|---|---|---|
| Model | Grammaticality | Importance |
| TRANSDUCER | 4.06$^{\dagger*}$ | 3.21$^{\dagger*}$ |
| McDONALD | 3.05$^\dagger$ | 2.82$^\dagger$ |
| Reference | 4.52 | 3.70 |

| Ziff-Davis | | |
|---|---|---|
| Model | Grammaticality | Importance |
| TRANSDUCER | 4.07$^\dagger$ | 3.23$^\dagger$ |
| McDONALD | 3.98$^\dagger$ | 3.22$^\dagger$ |
| Reference | 4.65 | 4.12 |

Table 8: Mean ratings on compression output elicited by humans (*: sig. diff. from McDONALD ($\alpha < 0.01$); $^\dagger$: sig. diff. from Reference ($\alpha < 0.01$); using post-hoc Tukey tests)

worse performance than the Reference, save one exception. On the CLspoken corpus, there is no significant difference between the TRANSDUCER and the gold standard.

These results indicate that our highly expressive framework is a good model for sentence compression. Under several experimental conditions, across different domains, we obtain better performance than previous work. Importantly, the model described here is not compression-specific, it could be easily adapted to other tasks, corpora or languages (for which syntactic analysis tools are available). Being supervised, the model learns to fit the compression rate of the training data. In this sense, it is somewhat inflexible as it cannot easily adapt to a specific rate given by a user or imposed by an application (e.g., when displaying text on small screens). Nevertheless, compression rate can be indirectly manipulated by adopting loss functions that encourage or discourage compression or directly during decoding by stratifying the chart for length (McDonald, 2006).

## 8. Conclusions

In this paper we have formulated sentence compression as a tree-to-tree rewriting task.[19] We developed a system that licenses the space of all possible rewrites using a tree substitution grammar. Each grammar rule is assigned a weight which is learned discriminatively within a large margin model (Tsochantaridis et al., 2005). A specialized algorithm is used to learn the model weights and find the best scoring compression under the model. We argue that the proposed framework is appealing for several reasons. The synchronous grammar provides expressive power to capture rewrite operations that go beyond word deletion such as reordering, changes in non-terminal categories and lexical substitution. Since it is not deletion-specific, the model could be ported to other rewriting tasks (see Cohn & Lapata, 2008, for an example) without the overhead of devising new algorithms for decoding or training. Moreover, the discriminative nature of the learning algorithm allows for the incorporation of all manner of powerful features. The rich feature space in conjunction with the choice of an appropriate loss function afford greater flexibility in fitting the empirical data for different domains or tasks.

---

19. The source code is freely available from `http://homepages.inf.ed.ac.uk/tcohn/t3`.

We evaluated our model on three compression corpora (CLspoken, CLwritten, and Ziff-Davis) and showed that in most cases it yields results superior to state-of-the-art (McDonald, 2006). Our experiments were also designed to assess several aspects of the proposed framework such as the complexity of the synchronous grammar, the choice of loss function, the effect of various features, and the quality of the generated tree output. We observed performance improvements by allowing maximally general grammar rules to be specified once, producing larger and more lexicalized rules. This concurs with Galley and McKeown (2007) who also find that lexicalization yields better compression output. The choice of loss function appears to have less of an effect. We devised three classes of loss functions based on Hamming distance, Edit distance and F1 score. Overall, the simple token-based Hamming loss achieved the best results. We conjecture that this is due to its simplicity – it can be evaluated more precisely than many of the other loss functions and isn't affected by poor parser output. Our feature ablation study revealed that ngram features are beneficial, mirroring a similar finding in the machine translation literature (Chiang, 2007). Finally, we found that the trees created by our generation algorithm are more accurate compared to the output of a parser applied to the string output. This augurs well for use in a cascaded NLP pipeline, where other systems use the compression output as input for further processing, and can potentially make better use of the system output.

Future extensions are many and varied. An obvious extension concerns porting the framework to other rewriting applications such as document summarization (Daumé III & Marcu, 2002) or machine translation (Chiang, 2007). Initial work (Cohn & Lapata, 2008) shows that the tree-to-tree transduction model presented here can be easily adapted to a sentence abstraction task where compression takes place using rewrite operations that are not restricted to word deletion. Examples include substitution, reordering, and insertion. Other future directions involve more detailed feature engineering, including source conditioned features and ngram features besides the language model. More research is needed to establish suitable loss functions for compression and other rewriting tasks. In particular it should be interesting to experiment with loss functions that incorporate a wider range of linguistic features beyond parts of speech. Examples include losses based on parse trees and semantic similarity. Finally, the experiments presented in this work use a grammar acquired from the training corpus. However, there is nothing inherent in our formalization that restricts us to this particular grammar. We therefore plan to investigate the potential of our method with unsupervised or semi-supervised grammar induction techniques for other rewriting tasks including paraphrase generation and machine translation.

## Acknowledgments

We are grateful to Philip Blunsom for insightful comments and suggestions and to the anonymous referees whose feedback helped to substantially improve the present paper. Special thanks to James Clarke for sharing his implementations of Clarke and Lapata's (2008) and McDonald's (2006) models with us. We acknowledge the support of EPSRC (grants GR/T04540/01 and GR/T04557/01). This work has made use of the resources provided by the Edinburgh Compute and Data Facility (ECDF). The ECDF is partially supported by the eDIKT initiative. A preliminary version of this work was published in the proceedings of EMNLP/CoNLL 2007.

## References

Aho, A. V., & Ullman, J. D. (1969). Syntax directed translations and the pushdown assembler. *Journal of Computer and System Sciences*, *3*, 37–56.

Alshawi, H., Bangalore, S., & Douglas, S. (2000). Learning dependency translation models as collections of finite state head transducers. *Computational Linguistics*, *26*(1), 45–60.

Berger, A. L., Pietra, S. A. D., & Pietra, V. J. D. (1996). A maximum entropy approach to natural language processing. *Computational Linguistics*, *22*(1), 39–71.

Bikel, D. (2002). Design of a multi-lingual, parallel-processing statistical parsing engine. In *Proceedings of the 2nd International Conference on Human Language Technology Research*, pp. 24–27, San Diego, CA.

Briscoe, E. J., & Carroll, J. (2002). Robust accurate statistical annotation of general text. In *Proceedings of the Third International Conference on Language Resources and Evaluation*, pp. 1499–1504, Las Palmas, Gran Canaria.

Carroll, J., Minnen, G., Pearce, D., Canning, Y., Devlin, S., & Tait, J. (1999). Simplifying text for language impaired readers. In *Proceedings of the 9th Conference of the European Chapter of the Association for Computational Linguistics*, pp. 269–270, Bergen, Norway.

Chandrasekar, R., & Srinivas, C. D. B. (1996). Motivations and methods for text simplification. In *Proceedings of the 16th International Conference on Computational Linguistics*, pp. 1041–1044, Copenhagen, Danemark.

Chiang, D. (2007). Hierarchical phrase-based translation. *Computational Linguistics*, *33*(2), 201–228.

Clarke, J., & Lapata, M. (2006). Models for sentence compression: A comparison across domains, training requirements and evaluation measures. In *Proceedings of the 21st International Conference on Computational Linguistics and 44th Annual Meeting of the Association for Computational Linguistics*, pp. 377–384, Sydney, Australia.

Clarke, J., & Lapata, M. (2008). Global inference for sentence compression: An integer linear programming approach. *Journal of Artificial Intelligence Research*, *31*, 399–429.

Cohn, T., & Lapata, M. (2008). Sentence compression beyond word deletion. In *Proceedings of the 22nd International Conference on Computational Linguistics*, pp. 137–144, Manchester, UK.

Collins, M. (2002). Discriminative training methods for hidden Markov models: theory and experiments with perceptron algorithms. In *Proceedings of the 2002 Conference on Empirical Methods in Natural Language Processing*, pp. 1–8, Morristown, NJ.

Collins, M. J. (1999). *Head-driven statistical models for natural language parsing*. Ph.D. thesis, University of Pennsylvania, Philadelphia, PA.

Crammer, K., & Singer, Y. (2003). Ultraconservative online algorithms for multiclass problems. *Machine Learning*, *3*, 951–999.

Daumé III, H., & Marcu, D. (2002). A noisy-channel model for document compression. In *Proceedings of the 40th Annual Meeting of thev Association for Computational Linguistics*, pp. 449–456, Philadelphia, PA.

Eisner, J. (2003). Learning non-isomorphic tree mappings for machine translation. In *The Companion Volume to the Proceedings of 41st Annual Meeting of the Association for Computational Linguistics*, pp. 205–208, Sapporo, Japan.

Galley, M., Hopkins, M., Knight, K., & Marcu, D. (2004). What's in a translation rule?. In *Proceedings of the 2004 Human Language Technology Conference of the North American Chapter of the Association for Computational Linguistics*, pp. 273–280, Boston, MA.

Galley, M., & McKeown, K. (2007). Lexicalized Markov grammars for sentence compression. In *Proceedings of Human Language Technologies 2007: The Conference of the North American Chapter of the Association for Computational Linguistics*, pp. 180–187, Rochester, NY.

Grael, J., & Knight, K. (2004). Training tree transducers. In *Proceedings of the 2004 Human Language Technology Conference of the North American Chapter of the Association for Computational Linguistics*, pp. 105–112, Boston, MA.

Hermjakob, U., Echihabi, A., & Marcu, D. (2002). Natural language based reformulation resource and wide exploitation for question answering. In *Proceedings of 11th Text Retrieval Conference*, Gaithersburg, MD.

Hori, C., & Furui, S. (2004). Speech summarization: an approach through word extraction and a method for evaluation. *IEICE Transactions on Information and Systems*, *E87-D*(1), 15–25.

Jing, H. (2000). Sentence reduction for automatic text summarization. In *Proceedings of the 6th Applied Natural Language Processing Conference*, pp. 310–315, Seattle, WA.

Joachims, T. (2005). A support vector method for multivariate performance measures. In *Proceedings of the 22nd International Conference on Machine Learning*, pp. 377–384, Bonn, Germany.

Kaji, N., Okamoto, M., & Kurohashi, S. (2004). Paraphrasing predicates from written language to spoken language using the web. In *Proceedings of the 2004 Human Language Technology Conference of the North American Chapter of the Association for Computational Linguistics*, pp. 241–248, Boston, MA.

Knight, K. (1999). Decoding complexity in word-replacement translation models. *Computational Linguistics*, *25*(4), 607–615.

Knight, K., & Marcu, D. (2002). Summarization beyond sentence extraction: a probabilistic approach to sentence compression. *Artificial Intelligence*, *139*(1), 91–107.

Lin, D., & Pantel, P. (2001). Discovery of inference rules for question answering. *Natural Language Engineering*, *7*(4), 342–360.

McDonald, R. (2006). Discriminative sentence compression with soft syntactic constraints. In *Proceedings of the 11th Conference of the European Chapter of the Association for Computational Linguistics*, pp. 297–304, Trento, Italy.

Nguyen, M. L., Horiguchi, S., Shimazu, A., & Ho, B. T. (2004). Example-based sentence reduction using the hidden markov model. *ACM Transactions on Asian Language Information Processing*, *3*(2), 146–158.

Och, F. J., & Ney, H. (2004). The alignment template approach to statistical machine translation. *Computational Linguistics*, *30*(4), 417–449.

Papineni, K., Roukos, S., Ward, T., & Zhu, W.-J. (2002). BLEU: a method for automatic evaluation of machine translation. In *Proceedings of the 40th Annual Meeting of thev Association for Computational Linguistics*, pp. 311–318, Philadelphia, PA.

Petrov, S., Barrett, L., Thibaux, R., & Klein, D. (2006). Learning accurate, compact, and interpretable tree annotation. In *Proceedings of the 21st International Conference on Computational Linguistics and 44th Annual Meeting of the Association for Computational Linguistics*, pp. 433–440, Sydney, Australia.

Riezler, S., King, T. H., Crouch, R., & Zaenen, A. (2003). Statistical sentence condensation using ambiguity packing and stochastic disambiguation methods for lexical-functional grammar. In *Proceedings of the 2003 Human Language Technology Conference of the North American Chapter of the Association for Computational Linguistics*, pp. 118–125, Edmonton, Canada.

Shieber, S., & Schabes, Y. (1990). Synchronous tree-adjoining grammars. In *Proceedings of the 13th International Conference on Computational Linguistics*, pp. 253–258, Helsinki, Finland.

Stolcke, A. (2002). SRILM – an extensible language modeling toolkit. In *Proceedings of the International Conference on Spoken Language Processing*, Denver, CO.

Tsochantaridis, I., Joachims, T., Hofmann, T., & Altun, Y. (2005). Large margin methods for structured and interdependent output variables. *Journal of Machine Learning Research*, *6*, 1453–1484.

Turner, J., & Charniak, E. (2005). Supervised and unsupervised learning for sentence compression. In *Proceedings of the 43rd Annual Meeting of the Association for Computational Linguistics*, pp. 290–297, Ann Arbor, MI.

Vandeghinste, V., & Pan, Y. (2004). Sentence compression for automated subtitling: A hybrid approach. In *Text Summarization Branches Out: Proceedings of the ACL-04 Workshop*, pp. 89–95, Barcelona, Spain.

Wu, D. (1997). Stochastic inversion transduction grammars and bilingual parsing of parallel corpora. *Computational Linguistics, 23*(3), 377–404.

Yamada, K., & Knight, K. (2001). A syntax-based statistical translation model. In *Proceedings of the 39th Annual Meeting of the Association for Computational Linguistics*, pp. 523–530, Toulouse, France.